\documentclass[10pt,twocolumn,letterpaper]{article}

\usepackage{cvpr}
\usepackage{times}
\usepackage{epsfig}
\usepackage{graphicx}
\usepackage{amsmath}
\usepackage{amssymb}
\usepackage{amsthm}

\usepackage[pagebackref=true,breaklinks=true,letterpaper=true,colorlinks,bookmarks=false]{hyperref}
\usepackage[numbers]{natbib}

\cvprfinalcopy 

\def\cvprPaperID{3461} 

\ifcvprfinal\fi

\begin{document}

\title{On the Effects of Batch and Weight Normalization in Generative Adversarial Networks}

\author{
  Sitao Xiang\textsuperscript{1}\hspace{4em}Hao Li\textsuperscript{1, 2, 3}\\
  \textsuperscript{1}University of Southern California\\
  \textsuperscript{2}Pinscreen\\
  \textsuperscript{3}USC Institute for Creative Technologies\\
  {\tt\small sitaoxia@usc.edu \hspace{2em} hao@hao-li.com} \\
}

\maketitle

\begin{abstract}

Generative adversarial networks (GANs) are highly effective unsupervised learning frameworks that can generate very sharp data, even for data such as images with complex, highly multimodal distributions.  However GANs are known to be very hard to train, suffering from problems such as mode collapse and disturbing visual artifacts. Batch normalization (BN) techniques have been introduced to address the training. Though BN accelerates the training in the beginning, our experiments show that the use of BN can be unstable and negatively impact the quality of the trained model. The evaluation of BN and numerous other recent schemes for improving GAN training is hindered by the lack of an effective objective quality measure for GAN models. To address these issues, we first introduce a weight normalization (WN) approach for GAN training that significantly improves the stability, efficiency and the quality of the generated samples. To allow a methodical evaluation, we introduce squared Euclidean reconstruction error on a test set as a new objective measure, to assess training performance in terms of speed, stability, and quality of generated samples. Our experiments with a standard DCGAN architecture on commonly used datasets (CelebA, LSUN bedroom, and CIFAR-10) indicate that training using WN is generally superior to BN for GANs, achiving 10\% lower mean squared loss for reconstruction and significantly better qualitative results than BN. We further demonstrate the stability of WN on a 21-layer ResNet trained with the CelebA data set.

\end{abstract}

\section{Introduction}

Despite their prevalent use, the effects of Batch Normalization (BN) \cite{ioffe2015batch} in Generative Adversarial Networks (GAN) \citep{goodfellow2014generative} have not been examined carefully. Popularized by  the influential DCGAN architecture \citep{radford2015unsupervised}, the use of BN in GANs is typically justified by its perceived training speedup and stability, but the generated samples often suffer from visual artifacts and limited variations (mode collapse). The lack of evidence that BN always improves GAN training is partly due to the unavailability of quality measures for GAN models.
Being puzzled by this technique, we propose a methodical evaluation of GAN models and assess their abilities to generate large variations of samples (mode coverage). The idea is to hold out a portion of the dataset as a test dataset and try to find the latent code that generates the closest approximation to these test images. For each test image, we optimize for the latent code by gradient descent for a fixed number of iterations. The average squared Euclidean distance between the test samples and the reconstructed ones is used as a measure of the quality of GANs.

Our experiments show that the reconstruction error correlates with the visual quality of the generated samples, and while still time consuming, this approach is more efficient than existing log-likelihood-based evaluation methods. Our evaluation technique is therefore convenient for monitoring the progress during training.
We show that BN generally accelerates training in early stages, and can increase the success rate of GAN training for certain datasets and network structures where a model without any normalization could often fail. In many cases though, BN can cause the stability and generalization power of the model to decrease drastically.
Following the work of Salimans and Kingma \citep{salimans2016weight} and Arpit \etal \citep{arpit2016normalization}, we introduce a modified Weight Normalization (WN) technique for GAN training. Using the same sets of experiments, we found that our WN approach can achieve faster and more stable training than BN, as well as generate equal or higher quality samples than GAN models without normalization. We believe that our proposed WN technique is superior than BN in the context of GANs.

\section{Related Work}

\paragraph{Batch Normalization.}

Batch Normalization (BN) \citep{ioffe2015batch} is a technique to accelerate the training of deep neural networks and has been shown to be effective in various applications. In the context of GANs, it first appeared in LAPGAN by Denton \etal \citep{denton2015deep} (for generator only), and made popular by the influential DCGAN architecture by Radford \etal \citep{radford2015unsupervised} (for both generator and discriminator). It has since become a common practice, as listed in this overview of GAN techniques \citep{ganhack} and used in many GAN architectures (e.g. WGAN \citep{arjovsky2017wasserstein} and EBGAN \citep{zhao2016energy}).
To summarize, BN takes a batch of samples $\{x_1,x_2,\ldots,x_m\}$ and computes the following:

\begin{equation}
y_i=\frac{x_i-\mu_{\mathcal{B}}}{\sigma_{\mathcal{B}}}\cdot\gamma+\beta \quad,
\end{equation}

where $\mu_{\mathcal{B}}$ and $\sigma_{\mathcal{B}}$ are the means and standard deviations of the input batch and $\gamma$ and $\beta$ are the learned parameters. As a result, the output will always have a mean $\beta$ and a standard deviation $\gamma$, regardless of the input distribution. Most importantly, the gradients must be back-propagated through the computation of $\mu_{\mathcal{B}}$ and $\sigma_{\mathcal{B}}$.

\paragraph{Weight Normalization.}

Weight Normalization (WN) is a more recent normalization technique proposed by Salimans and Kingma \citep{salimans2016weight}.
For a linear layer
\begin{equation}
\label{eqn:weightnorm1}
\mathbf{y}=\mathbf{W}^T\mathbf{x}+\mathbf{b}\quad,
\end{equation}
where $\mathbf{x}\in\mathbb{R}^n$, $\mathbf{y}\in\mathbb{R}^m$, $\mathbf{W}\in\mathbb{R}^{n\times m}$ and $\mathbf{b}\in\mathbb{R}^m$, weight normalization performs a reparameterization $\mathbf{W}$ with $\mathbf{V}\in\mathbb{R}^{n\times m}$ and $\mathbf{g}\in\mathbb{R}^m$:
\begin{equation}
\label{eqn:weightnorm2}
\mathbf{w}_i=\frac{g_i}{||\mathbf{v}_i||_2}\cdot\mathbf{v}_i\quad,
\end{equation}
where $\mathbf{w}_i$ and $\mathbf{v}_i$ are the $i$-th column of $\mathbf{W}$ and $\mathbf{V}$, respectively. As with BN, the computation of $||\mathbf{v}_i||_2$ is taken into account when computing the gradient with respect to $\mathbf{V}$.

Although presented as a reparameterization that modifies the curvature of the loss function, the main idea is to simply divide the weight vectors by their norms. A very similar idea has been proposed around the same time, under ``normalization propogation'' (NormProp) by Arpit \etal \citep{arpit2016normalization}. While the effectiveness of this technique has been illustrated on various experiments in \citep{salimans2016weight} and \citep{arpit2016normalization}, they did not investigate this acceleration approach for GANs. As detailed in Section \ref{sec:weightnorm}, we propose a modified version of Weight Normalization to improve the training of GAN models.
 
\paragraph{GAN Evaluation.}

In earlier GAN-related works, with a lack of quantitative measures, visual inspection has been a commonly used method. In addition to inspecting visual quality, this has also been used to show that the model did not overfit, by interpolation in latent space (e.g. \citep{denton2015deep}) and by finding closest training sample to generated samples and point out their difference (e.g. \citep{goodfellow2014generative}).

Various quantitative measures has since been proposed. A commonly used one is estimating the log-likelihood of the training set in the generator's distribution, by generating a large amount of samples and fitting a Gaussian Parzen window (e.g. \citep{denton2015deep, makhzani2015adversarial}). As discussed by Theis \etal \citep{theis2015note}, this is not particularly effective, as the amount of samples that need to be generated for accurate log-likelihood estimation is intractable.

Another measure is Inception score, proposed by Salimans \etal \citep{salimans2016improved}, based on the assumption that a good generative model should be able to generate meaningful objects. A limitation of this approach is that, the inception model is pretrained on another image classification task, usually for natural objects. Thus, it is only useful as a measure of GAN quality trained on images on similar objects.
The quality of GANs has also been evaluated indirectly, e.g. by measuring the classification accuracy using features extracted by a GAN discriminator \citep{radford2015unsupervised}.
Our proposed measure of reconstruction loss is most similar to that used by Metz \etal \citep{metz2016unrolled}. We discuss differences in section \ref{sec:evaluation}.

\section{Weight Normalization for GAN Training}
\label{sec:weightnorm}

We propose a modified formulation of the weight normalization approach introduced by Salimans and Kingma \citep{salimans2016weight}.
A notable deficiency of the original WN technique is that, in its simplest form, it does not normalize the mean value of the input. In \citep{salimans2016weight} this is solved by augmenting WN with a version of BN that only normalizes the mean of the input but not the variance. While their experiments showed an improved performance for the CIFAR-10 classification task compared to plain WN, it gave worse results for several of our experiments (See appendix \ref{sec:moremodels}). Hence, we chose to not include this augmentation in our approach.

In \citep{arpit2016normalization}, the authors attempt to solve this problem by enforcing a zero-mean, unit-variance distribution throughout the network. In their method, the scale and bias are first fixed as $\mathbf{g}=\mathbf{1}$ and $\mathbf{b}=\mathbf{0}$, that is,

\begin{equation}
\label{eqn:strictweightnorm}
y=\frac{\mathbf{w}^T\mathbf{x}}{||\mathbf{w}||_2}\quad.
\end{equation}

For simplicity, we consider here a single output neuron. The training data is normalized so that the input to the network has zero mean and unit variance. The mean and variance of the output of each nonlinear layer (ReLU in this case) is evaluated in closed form under the assumption that the input to the preceding linear layer is from a multivariate standard normal distribution. The mean and variance is then used to correct the distribution of the output:

\begin{equation}
\label{eqn:normprop0}
y = \frac{1}{\sigma}\left[\mathrm{ReLU}\left(\frac{\mathbf{w}^T\mathbf{x}}{||\mathbf{w}||_2}\right)-\mu\right]\quad,
\end{equation}

where

\begin{equation}
\mu=\sqrt{\frac{1}{2\pi}} \quad \textrm{and} \quad \sigma=\sqrt{\frac{1}{2}\left(1-\frac{1}{\pi}\right)}
\end{equation}

are the mean and standard deviation of the distributions after ReLU when $\mathbf{x}$ is from a multivariate standard normal distribution. The output $y$ in equation \ref{eqn:normprop0} would also have zero mean and unit variance.

Notice that this ad-hoc fix does not really achieve its goal. Firstly, as mentioned in~\citep{arpit2016normalization}, the closed form mean and variance is only an approximation since the correctness of this derivation requires the input to be normal distributed, which does not strictly hold beyond the first layer. 
More critically, after deriving equation \ref{eqn:normprop0} and fixing $\mu$ and $\sigma$, akin to Batch Normalization, they argue that an affine transformation needs to be learned after the weight-normalized linear layer and before the succeeding non-linear layer, in order to avoid decreasing the set of functions that can be represented by the network. 
The formulation then becomes:

\begin{equation}
\label{eqn:normprop1}
y = \frac{1}{\sigma}\left[\mathrm{ReLU}\left(\frac{\gamma\left(\mathbf{w}^T\mathbf{x}\right)}{||\mathbf{w}||_2}+\beta\right)-\mu\right]\quad.
\end{equation}

Their derivation for $\mu$ and $\sigma$ is for the restricted case, when there is no learned affine transformation, i.e. when $\gamma=1$ and $\beta=0$. When this restriction is relaxed, the result would be invalid, even if the i.i.d. normal condition on $\mathbf{x}$ does hold. We could make $\mu$ and $\sigma$ functions of $\gamma$ and $\beta$ to fix this error, but the back-propagation computation would be overly complex, since these functions also need to be taken into account.

As we cannot hope to strictly enforce a zero-mean unit-variance distribution, we propose to use a simpler approximation instead. Note that with ReLU-like nonlinearity (i.e. ReLU, leaky ReLU and parametric ReLU) we have $\mathrm{ReLU}(ax)=a\cdot\mathrm{ReLU}(x)$ when $a\ge 0$. In equation \ref{eqn:normprop1}, when $\gamma<0$, we can always invert the direction of $\mathbf{w}$ and take the negative of $\gamma$. Hence, without loss of generality, we can assume $\gamma\ge 0$. Then equivalently, equation \ref{eqn:normprop1} can be written as

\begin{equation}
\label{eqn:normprop2}
y = \frac{\gamma}{\sigma}\left[\mathrm{ReLU}\left(\frac{\mathbf{w}^T\mathbf{x}}{||\mathbf{w}||_2}+\frac{\beta}{\gamma}\right)-\frac{\mu}{\gamma}\right]\quad.
\end{equation}

The purpose of $\mu$ and $\sigma$ is to cancel out the mean and variance introduced by ReLU and the affine transformation (i.e. $\beta$ and $\gamma$). Instead of deriving a complex formula, we simply set $\mu=\beta$ and $\sigma=\gamma$, and re-formulate the equation using $\alpha=-\frac{\beta}{\gamma}=-\frac{\mu}{\gamma}$. Equation \ref{eqn:normprop2} becomes:

\begin{equation}
\label{eqn:normprop3}
y=\mathrm{ReLU}\left(\frac{\mathbf{w}^T\mathbf{x}}{||\mathbf{w}||_2}-\alpha\right)+\alpha\quad.
\end{equation}

Note that we can now separate out the restricted weight normalized layer from equation \ref{eqn:normprop3}. We call the remaining part ``Translated ReLU (TReLU)'':

\begin{align}
\mathrm{TReLU}_\alpha(x)=&\mathrm{ReLU}(x-\alpha)+\alpha\\
=&\begin{cases}
x & (x\ge\alpha)\\
\alpha & (x<\alpha)\quad,
\end{cases}
\end{align}

where $\alpha$ is a learned parameter. It is more commonly referred to as a ``threshold layer'', defined by $y=\max\{x,\alpha\}$, but here the threshold is learned. We chose this name to reflect the fact that other ReLU-like nonlinear functions can be used to give translated leaky and parametric ReLU layers. Here, we ``translate'' the data by $-\alpha$, apply the nonlinear function, then ``translate'' the data ``back'' (by $\alpha$). By using TReLU instead of adding bias to the previous layer, we prevent (to a certain degree) the introduction of a large mean into the distribution.

This simplification effectively negates the learned affine transformation, which seemingly would reduce the set of functions that can be represented by the network. We argue, however, that allowing the learning of an affine transformation at the last weight-normalized layer recovers the expressiveness of the entire stack of layers (see appendix \ref{sec:equivalence} for proof). From now on, ``strict weight-normalized layers'' will refer to layers without affine transformations (Equation \ref{eqn:strictweightnorm}), while layers with a learned affine transformation

\begin{equation}
y=\frac{\mathbf{w}^T\mathbf{x}}{||\mathbf{w}||_2}\cdot\gamma+\beta
\end{equation}

are referred as ``affine weight-normalized layers''. These are collectively called ``weight-normalized layers''.

\section{Evaluation Method}
\label{sec:evaluation}

For many generative models, the reconstruction error on the training set is often explicitly optimized in some form (e.g., Variational Autoencoders \citep{kingma2013auto}). Even when this is not the case as in GANs, it is natural to evaluate the model with a reconstruction loss (squared Euclidean distance) measured on a test set. In the case of GANs, given a generator $G$ and a set of test samples $X=\{x^{(1)},x^{(2)},\ldots,x^{(m)}\}$, the reconstruction loss of $G$ on $X$ is defined as

\begin{equation}
\mathcal{L}_\mathrm{rec}(G,X)=\frac{1}{m}\sum_{i=1}^m\min_z||G(z)-x^{(i)}||_2^2\quad.
\end{equation}

In the case of images, we normalize for different image sizes by considering per pixel, per color channel reconstruction loss, thus we divide the loss by $3wh$ where $w$ and $h$ are the width and height of the training images. Since there is no way to directly infer the optimal $z$ from $x$, we use an alternative method: starting from an all-zero vector, we perform gradient descent on the latent code to find one that minimizes the squared Euclidean distance between the sample generated from the code and the target one. Because the code is optimized instead of computed from a feed-forward network, the evaluation process is time-consuming. Thus, we avoid performing this evaluation at every training iteration when monitoring the training process, and only use a reduced number of samples and gradient descent steps. Only for the final trained model, we perform an extensive evaluation on a larger test set, with a larger number of steps.

This method is very similar to that proposed by Metz \etal \citep{metz2016unrolled}. There are two important differences: in \citep{metz2016unrolled} the samples used for reconstruction come from the training set, while we take the samples from a separate test set. Intuitively, in order to generate the test samples that are not in the training set, the generator must learn the distribution of the training samples, but not memorize and overfit on them. Such an effect would not be achieved if the test samples come from the training set.

Furthermore, \citep{metz2016unrolled} uses L-BFGS for optimization on the latent code. L-BFGS is known to give good and fast optimization for problems that are not too high-dimension, which suits the setting of this problem well. However, its effectiveness is sensitive to many of its parameters. We were not able to find a combination of parameters that consistently work well under the various experiment settings. This also made it harder to justify the choice of parameters since for the different models we would like to compare the best parameters may be very different.

Instead we use RMSProp. It may not be the fastest optimization method for this problem, but we found it to work well under our settings, and altering the parameters (learning rate and number of steps) generally affect the reconstruction result of different models in the same way, which makes comparison easier.

\section{Experiments}
\label{sec:experiments}

We conducted experiments on image generation tasks, with quantitative analysis on DCGAN-based architecture on CelebA, LSUN bedroom and CIFAR-10 datasets, and qualitative results with a 21-layer ResNet on CelebA. The CelebA experiments are detailed here. Due to limited space, we only show some generated and reconstructed samples on LSUN and CIFAR-10 here, and discuss the settings, qualitative and quantitative results along with more samples in Appendices \ref{sec:cifar} and \ref{sec:lsun}.

\subsection{DCGAN Setup}
\label{sec:setup}

For CelebA \citep{liu2015faceattributes}, we using central 160 $\times$ 160 patches. We compared three DCGAN-based models: (1) trained without any normalization as a reference (the non-normalized or ``vanilla'' model), (2) with Batch Normalization (``BN model''), and (3) with our formulation of Weight Normalization (``WN model''). The network is structured in the following way: for the discriminator, we use successive convolution layers with kernel size 4, stride 2, padding 1 and output features doubling that of the previous layer, starting from 64 features in the first layer. We add convolution layers until the spatial size of the feature map is sufficiently small (5$\times$5). We then add one final convolution layer with stride 1, zero padding and kernel size 5 (equaling the size of the last feature map). For the generator, we reverse this structure and use transposed convolution layers.

As per common practice, Batch Normalization is not applied to the first layer of the discriminator, nor to the last layers of both the discriminator and generator. Weight normalization is used for every layer. For the last layer of both discriminator and generator, we use affine weight-normalized layers (AWNConv) while for every other layer we use strict weight-normalized layers (SWNConv). Parametric ReLU (PReLU) is used for vanilla and batch-normalized models and Translated Parametric ReLU (TPReLU) for weight-normalized models. Slope and bias parameters are learned per-channel. The length of the code is 256 for all models. The architectures are summarized in table \ref{tab:models}. 
Additional details regarding the implementation of weight normalized layers are discussed in appendix \ref{sec:detail}.

\begin{table}
\caption{\label{tab:models} Network structure for discriminators (top) and generators (bottom). First three columns: type of layers for vanilla, BN and WN models, respectively. Fourth column: kernel size, stride, padding and number of output channels of convolution layer.}
\small
\begin{center}
\begin{tabular}{cccc}\hline
vanilla & BN      & WN      &               \\\hline
Conv    & Conv    & SWNConv & 4, 2, 1, 64   \\
-       & -       & -       &               \\
PReLU   & PReLU   & TPReLU  &               \\
Conv    & Conv    & SWNConv & 4, 2, 1, 128  \\
-       & BN      & -       &               \\
PReLU   & PReLU   & TPReLU  &               \\
Conv    & Conv    & SWNConv & 4, 2, 1, 256  \\
-       & BN      & -       &               \\
PReLU   & PReLU   & TPReLU  &               \\
Conv    & Conv    & SWNConv & 4, 2, 1, 512  \\
-       & BN      & -       &               \\
PReLU   & PReLU   & TPReLU  &               \\
Conv    & Conv    & SWNConv & 4, 2, 1, 1024 \\
-       & BN      & -       &               \\
PReLU   & PReLU   & TPReLU  &               \\
Conv    & Conv    & AWNConv & 5, 1, 0, 1    \\
Sigmoid & Sigmoid & Sigmoid &               \\\hline
\end{tabular}
\begin{tabular}{cccc}\hline
vanilla & BN      & WN      &               \\\hline
Conv    & Conv    & SWNConv & 5, 1, 0, 1024 \\
-       & BN      & -       &               \\
PReLU   & PReLU   & TPReLU  &               \\
Conv    & Conv    & SWNConv & 4, 2, 1, 512  \\
-       & BN      & -       &               \\
PReLU   & PReLU   & TPReLU  &               \\
Conv    & Conv    & SWNConv & 4, 2, 1, 256  \\
-       & BN      & -       &               \\
PReLU   & PReLU   & TPReLU  &               \\
Conv    & Conv    & SWNConv & 4, 2, 1, 128  \\
-       & BN      & -       &               \\
PReLU   & PReLU   & TPReLU  &               \\
Conv    & Conv    & SWNConv & 4, 2, 1, 64   \\
-       & BN      & -       &               \\
PReLU   & PReLU   & TPReLU  &               \\
Conv    & Conv    & AWNConv & 4, 2, 1, 3    \\
Sigmoid & Sigmoid & Sigmoid &               \\\hline
\end{tabular}
\end{center}
\end{table}

All models are optimized with RMSProp \citep{tieleman2012lecture}, with a learning rate of $10^{-4}$, $\alpha=0.9$, $\varepsilon=10^{-6}$ and a batch size of 32. Specifically for the BN model, we use separate batches for true samples and generated samples when training the discriminator, as suggested by \citep{ganhack}.
After each parameter update, we clip the learned slope of parametric ReLU layers to $[0,1]$.
There are a total of 202,599 images in CelebA dataset. We randomly selected 2,000 images for evaluation and used the rest for training. During the training, we perform a ``running evaluation'' for every 500 training iterations, on a randomly selected and fixed subset of 200 test samples. The optimal code is found by performing gradient descent for 50 steps, starting from a zero vector. Again we use RMSProp, with a learning rate of 0.01.

For each model, the best performing network during the training is saved and used for final evaluation. In the final evaluation, we use all 2,000 test samples and perform gradient descent for 2,000 steps. For BN model, we use its inference mode. In addition, we also use the ``converged'' model for evaluation, in case the model does converge but gives notably worse running reconstruction than the optimal recorded model. However, this did not occur in the main experiment. We consider that training has converged if both the running reconstruction loss and the generated samples stay stable for a sufficient amount of time.

We observed mode collapse issues with both the vanilla and BN models. To reduce the possibility that these observations are caused by random factors, we repeat the training procedure for these models three times. We present the results from the best training instances and additional ones of the vanilla and BN models can be found in Appendix \ref{sec:moreinstances}.

\subsection{Reconstruction}

\begin{figure}
\begin{center}
\includegraphics[width=\linewidth]{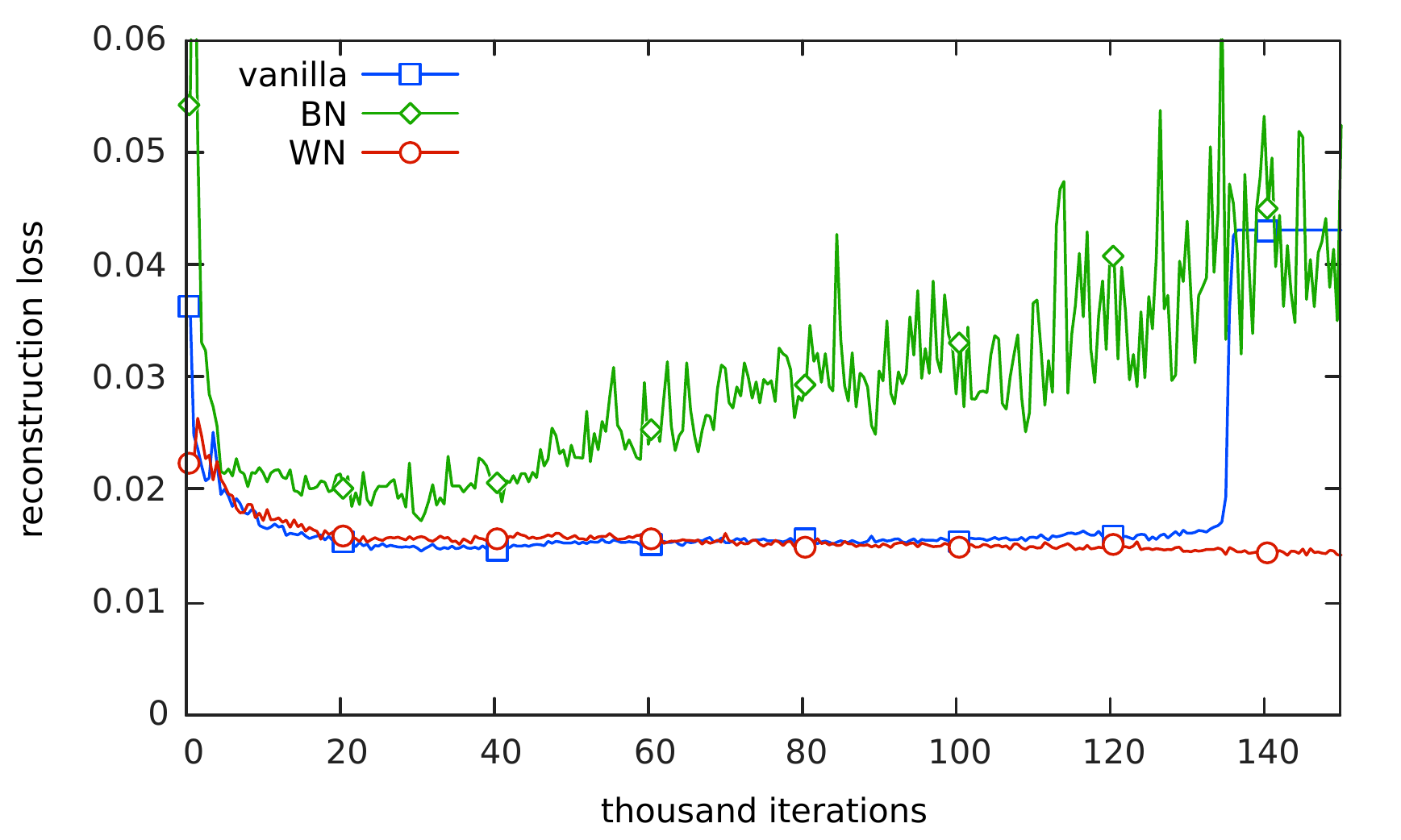}
\end{center}
\caption{\label{fig:recloss} Running reconstruction loss during training.}
\end{figure}

\begin{table}
\caption{\label{tab:recloss} Optimal reconstruction loss of the models}
\begin{center}
\begin{tabular}{cccc}\hline
Model   & Optimal iteration & Running loss & Final loss \\\hline
vanilla & 30,500            & 0.014509     & 0.006171        \\
BN      & 30,500            & 0.017199     & 0.006355        \\
WN      & 463,000           & 0.013010     & 0.005524        \\\hline
\end{tabular}
\end{center}
\end{table}

The running reconstruction loss of the three models is shown in Figure \ref{fig:recloss} for the first 150,000 iterations. The generated samples from both the vanilla and BN models have collapsed. The WN model was trained to 700,000 iterations and is considered to have converged (see Appendix \ref{sec:moresamples} for the prolonged training).

The lowest running reconstruction loss recorded during training, the iteration at which this minimum loss is achieved, and the final reconstruction loss for each model is listed in table \ref{tab:recloss}.
WN achieves about 10.5\% lower final reconstruction loss than the vanilla model, while for BN the loss is 3\% higher. We can also see from the loss curve that, until the vanilla model collapses, BN never achieved a better reconstruction loss.

\begin{figure}
\begin{center}
\includegraphics[width=\linewidth]{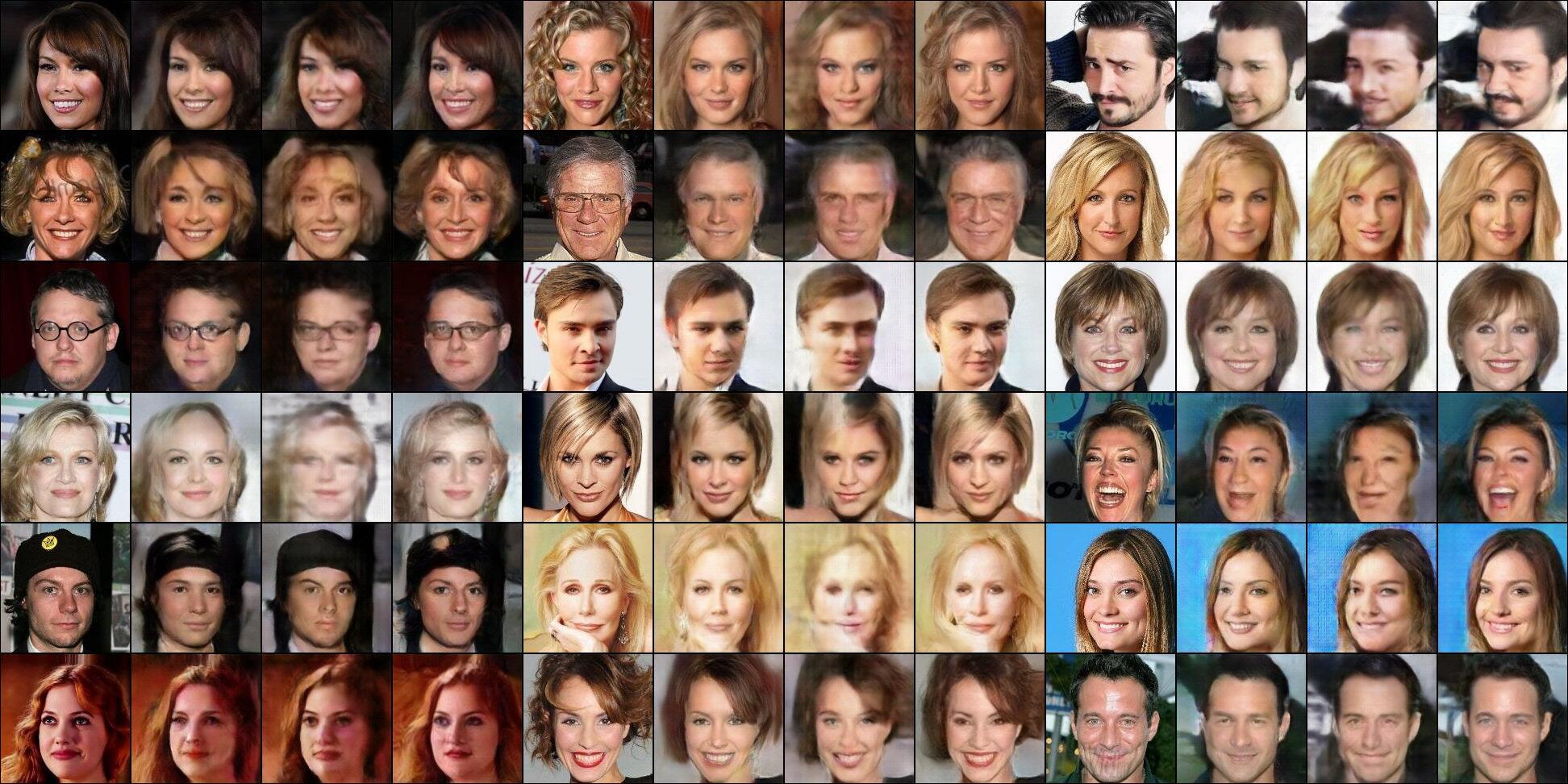}
\end{center}
\caption{\label{fig:recsample} Selected final reconstruction results. From left to right in each group: test sample, vanilla reconstruction, BN reconstruction, WN reconstruction. All images best viewed enlarged.}
\end{figure}

We also provide qualitative results of the reconstructions. Selected reconstructed samples are compared to the original test samples in figure \ref{fig:recsample}. These samples are selected such that all three models give reasonable results. Random samples can be found in Appendix \ref{sec:moresamples}. The WN model captures details (e.g. facial expression, texture of hair, subtle color variation) much more faithfully. Samples reconstructed by the BN model are significantly blurrier and affected by artifacts.

\subsection{Stability}

As shown in Figure \ref{fig:recloss}, the reconstruction of vanilla and BN models started to get worse relatively early on during their training, after achieving their optimal reconstruction loss. For the vanilla model, the loss went up slowly, then in a relatively short time around iteration 135,000, the generator collapses and produces the same output, which caused the reconstruction loss to increase suddenly. For the BN model, at around 40,000 iterations, the loss started to show excessive fluctuation. Our WN model however, kept improving steadily until 300,000 iterations and then remained largely stable.

\begin{figure}
\begin{center}
\includegraphics[width=\linewidth]{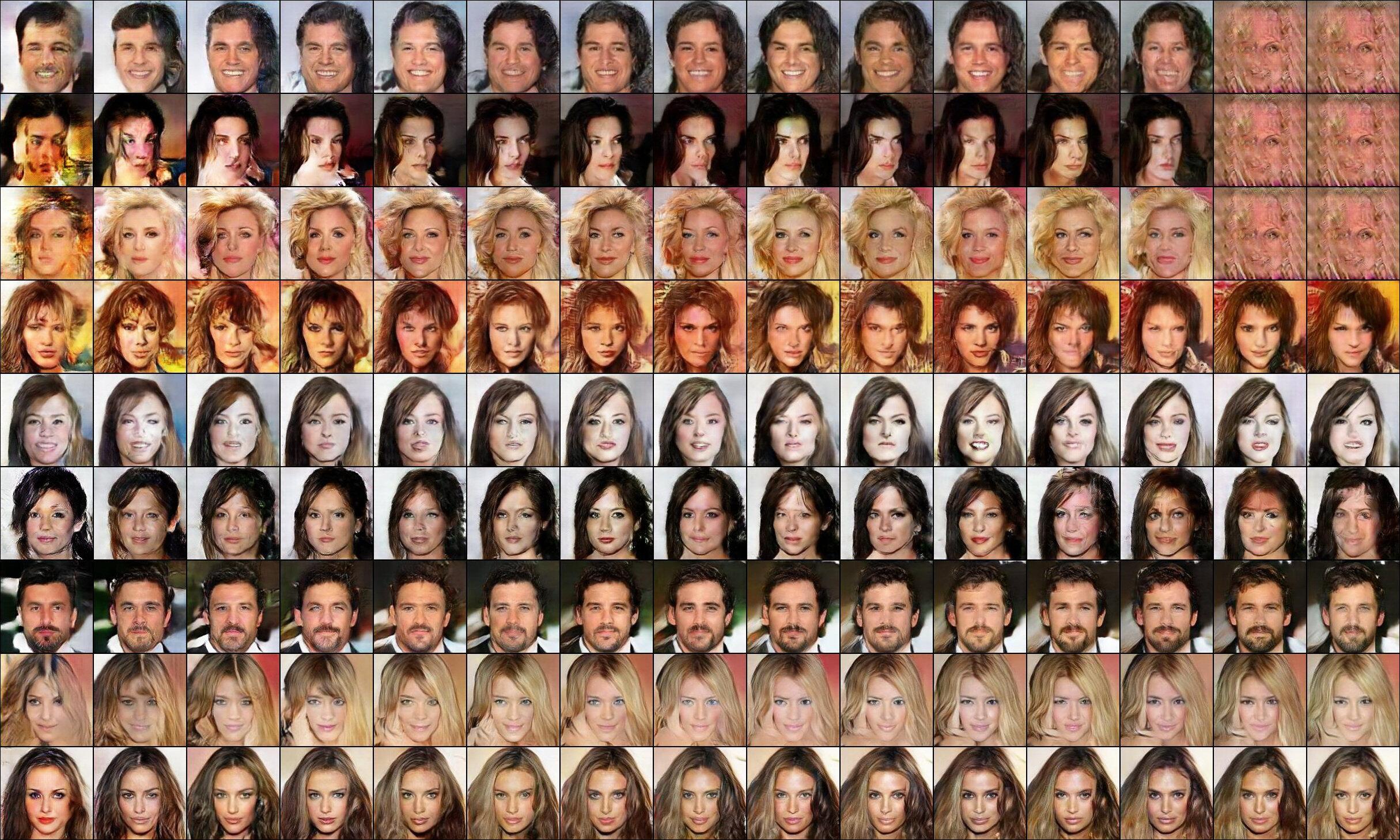}
\end{center}
\caption{\label{fig:stabilityvis} Evolution of samples during training. Top 3 rows: vanilla;  Middle 3 rows: BN; Bottom 3 rows: WN. Columns: every 10,000 iterations from 10,000 to 150,000.}
\end{figure}

We can also visualize this (in)stability by checking samples generated from the same code at different iterations, as shown in Figure \ref{fig:stabilityvis}. The WN model is noticeably more stable as samples generated from the same code remain mostly constant across a time scale of 100,000 iterations, and the generated samples are slowly improving, while the other two models produce more random variations. Additional visual analysis and samples can be found in Appendix \ref{sec:moresamples}.

\subsection{Training Speed}

\begin{figure}
\begin{center}
\includegraphics[width=\linewidth]{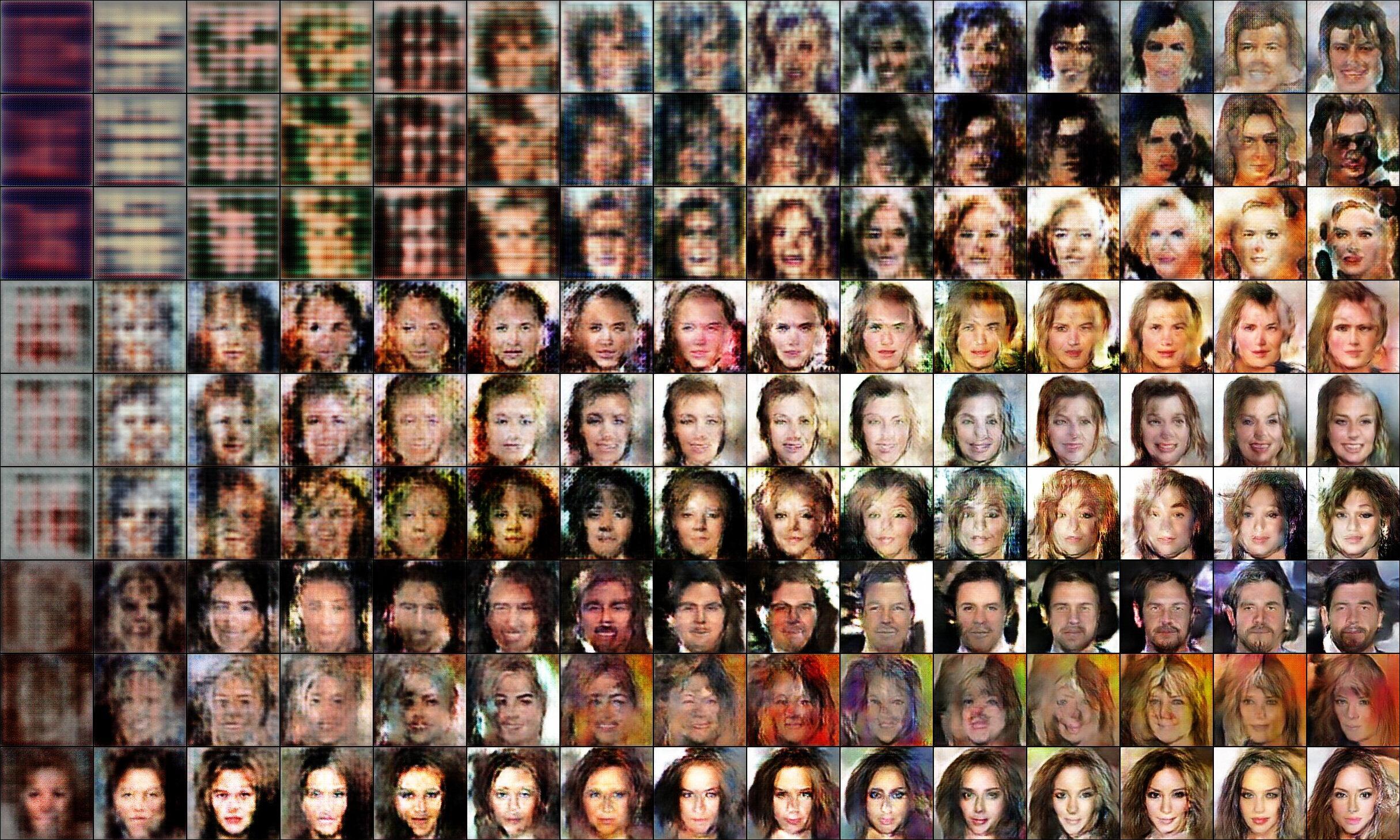}
\end{center}
\caption{\label{fig:speedvis} Evolution of samples during early stage of training. Top 3 rows: vanilla;  Middle 3 rows: BN; Bottom 3 rows: WN. Columns are samples from iterations 100, 200, 300, 400, 500, 600, 800, 1000, 1200, 1500, 2000, 2500, 3000, 4000 and 5000.}
\end{figure}

We compare the training speed of the three models by assessing their generated samples during early stages of the training, as illustrated in Figure \ref{fig:speedvis}. It is evident that Batch Normalization does accelerate training and the effect of Weight Normalization is comparable. Notice that our WN model can already produce a human face in only 100 iterations.
This accelerated training is mostly useful as a fast sanity check, when monitoring the training progress of deep neural networks. As shown in Figure, the visual quality of the samples generated by the three models are comparable at 10,000 iterations, and none of the models achieve a noticeably faster progression than the other. In addition, the ability to generate visually plausible samples earlier on does not necessarily translate into an overall faster improvement of the reconstruction.
Notice that BN allows a higher learning rate. The training of the vanilla and WN models often fail with a learning rate of $0.0002$, while the BN model can still be trainable with a learning rate of $0.001$. However, we found that an increased learning rate did not accelerate the training of the BN model. Instead, it further harms the stability of the model.

\subsection{Results on LSUN and CIFAR-10}

\begin{figure}
\begin{center}
\includegraphics[width=\linewidth]{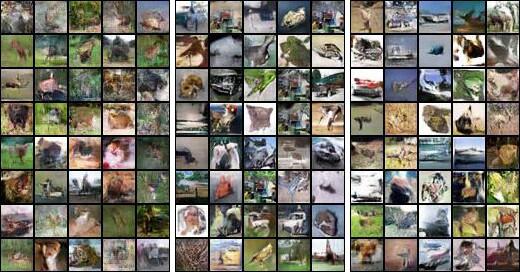}
\end{center}
\caption{\label{fig:cifar_samples_0} Random generated samples on CIFAR-10. Left to right: vanilla, BN, WN.}
\end{figure}

\begin{figure}
\begin{center}
\includegraphics[width=\linewidth]{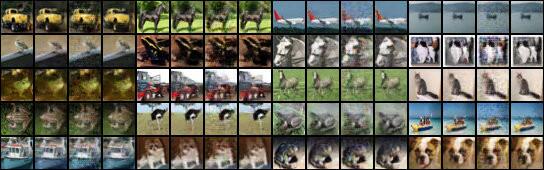}
\end{center}
\caption{\label{fig:cifar_rec_0} Random reconstructions on CIFAR-10. In each group, left to right: test sample, vanilla, BN, WN.}
\end{figure}

\begin{figure}
\begin{center}
\includegraphics[width=\linewidth]{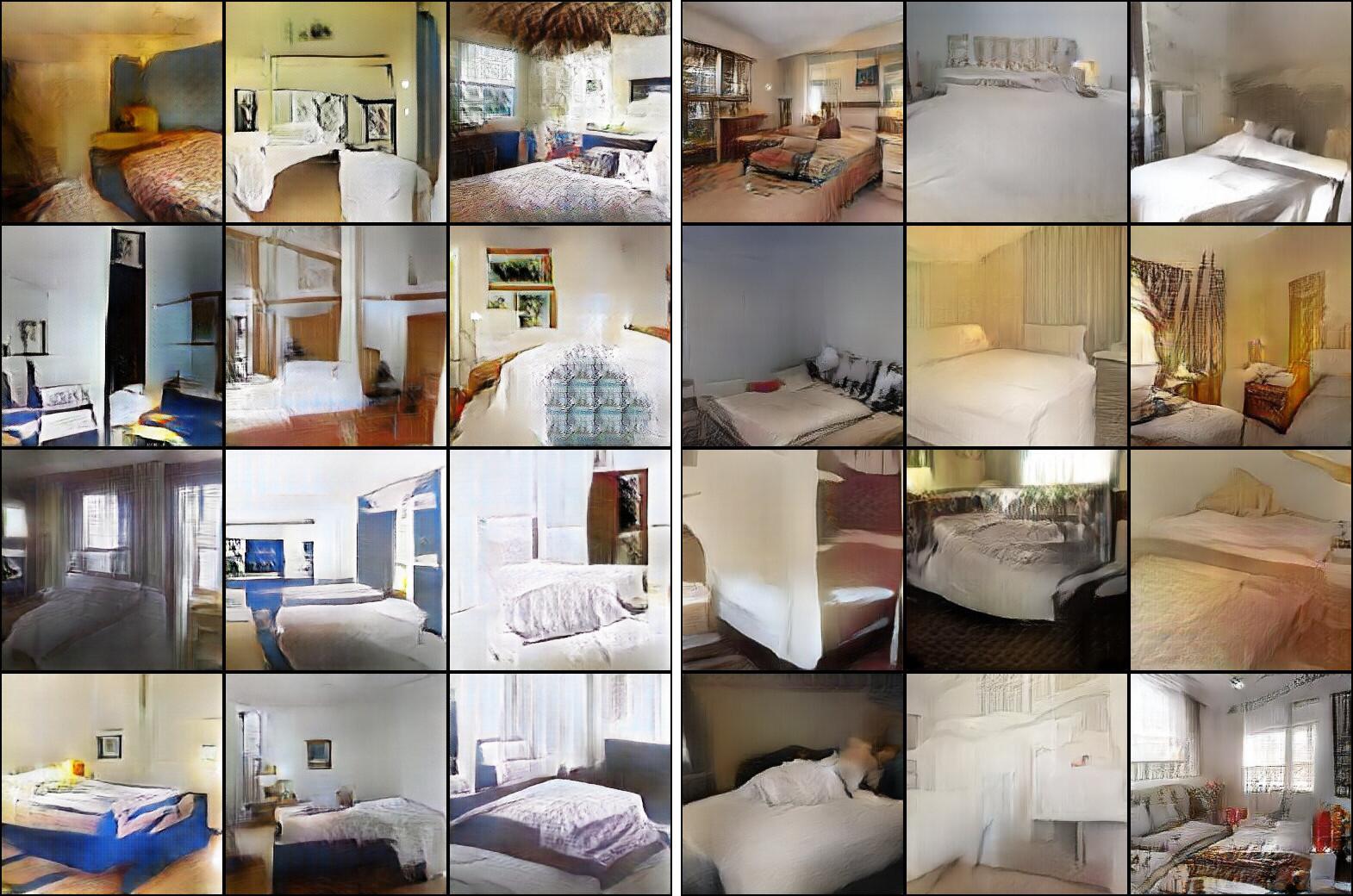}
\end{center}
\caption{\label{fig:lsun_samples_0} Random generated samples on LSUN. Left: BN, right: WN.}
\end{figure}

\begin{figure}
\begin{center}
\includegraphics[width=\linewidth]{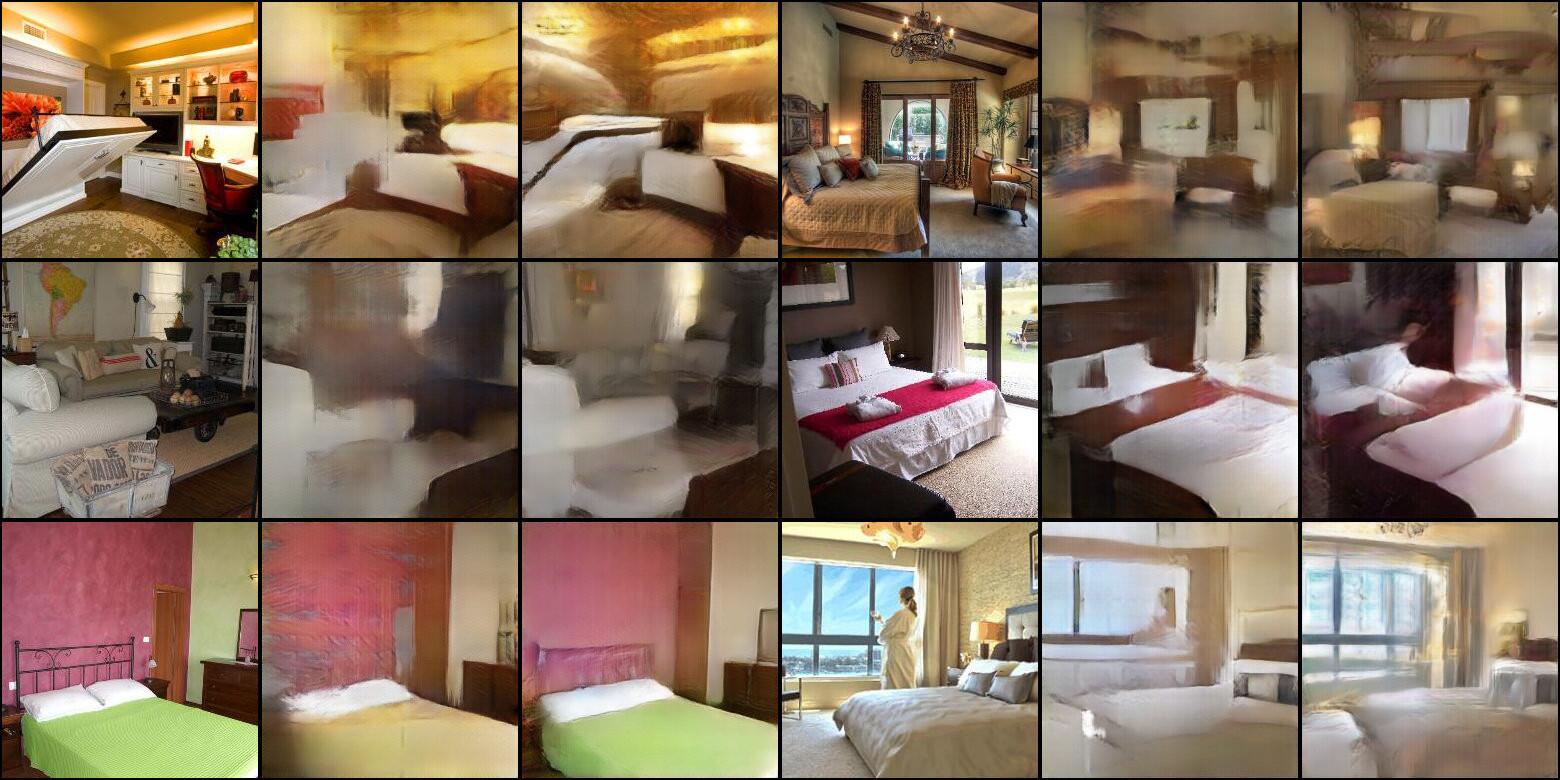}
\end{center}
\caption{\label{fig:lsun_rec_0} Random reconstructions on LSUN. In each group, left to right: test sample, BN, WN.}
\end{figure}

Figures \ref{fig:cifar_samples_0} through \ref{fig:lsun_rec_0} show random generated and reconstructed samples of the three models on CIFAR-10 and LSUN bedroom datasets. The vanilla model failed to train on LSUN, so only results for the BN and WN models are shown.

\subsection{ResNet Setup}

Residual Networks \citep{he2016deep} are becoming increasingly popular for image classification. While it has been used in GANs, the setting is usually an image-to-image translation task, e.g. image super-resolution \citep{ledig2016photo}. Direct image generation from noise with ResNet has not been particularly successful. Here we test our method on a 21-layer residual network.

Our block structure is as follows: we base our design on the basic blocks from \citep{he2016deep}. On the shortcut branch, we use an optional average pooling, present when the stride is 2, followed by an optional convolution with kernel size 1, present when the number of input features does not equal the number of output features. On the residue branch, we use Conv-BN-PReLU-Conv-BN structure for the BN model and remove batch normalization layers for the vanilla model. The two branches are then summed, then a final PReLU layer is applied to the result.

In the WN model, all convolutions are replaced with the strict weight normalized version and PReLU layers are replaced with the translated version. There is some complication when summing the two branches in the WN model, see Appendex \ref{sec:detail} for more details.
The first convolution on the residue branch has kernel size 3 or 4 when the stride is 1 or 2 respectively. the second convolution always have kernel size 3. The two convolutions always have padding 1. In the generator, all convolutions are replaced with transposed convolutions, and the average pooling on the shortcut branch is replaced with a nearest neighbour upscaling.

The discriminator network consists of 5 levels, with each level consisting of a stride 2 block followed by a stride 1 block with the same number of output features, for a total of 10 residue blocks and thus 20 layers. Then a final convolution with kernel size 5 and no padding is added, as in the DCGAN models above, for a total of 21 layers. Since the network is much deeper, to save computation time, we reduced the number of features to (64, 128, 256, 384, 512) and dimension of the latent space to 128. Again, the generator is a mirror image of the discriminator.
We also reduced the batch size to 16 and learning rate to $2\times 10^{-5}$.

\subsection{ResNet Results}

\begin{figure}
\begin{center}
\includegraphics[width=\linewidth]{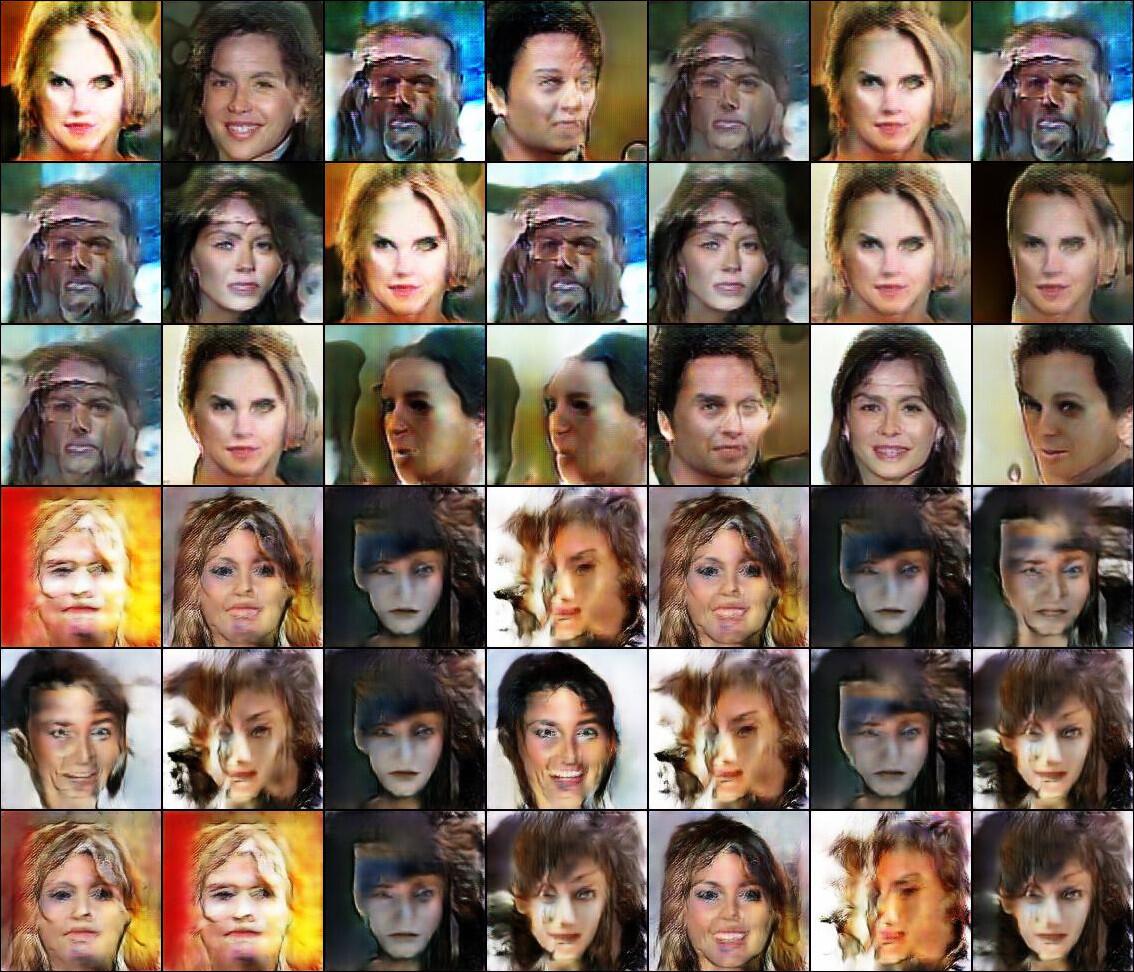}
\end{center}
\caption{\label{fig:resnet_samples} Random samples from vanilla and BN ResNet model, at iteration 70,000. Top 3 rows: vanilla; Bottom 3 rows: BN.}
\end{figure}

\begin{figure}
\begin{center}
\includegraphics[width=\linewidth]{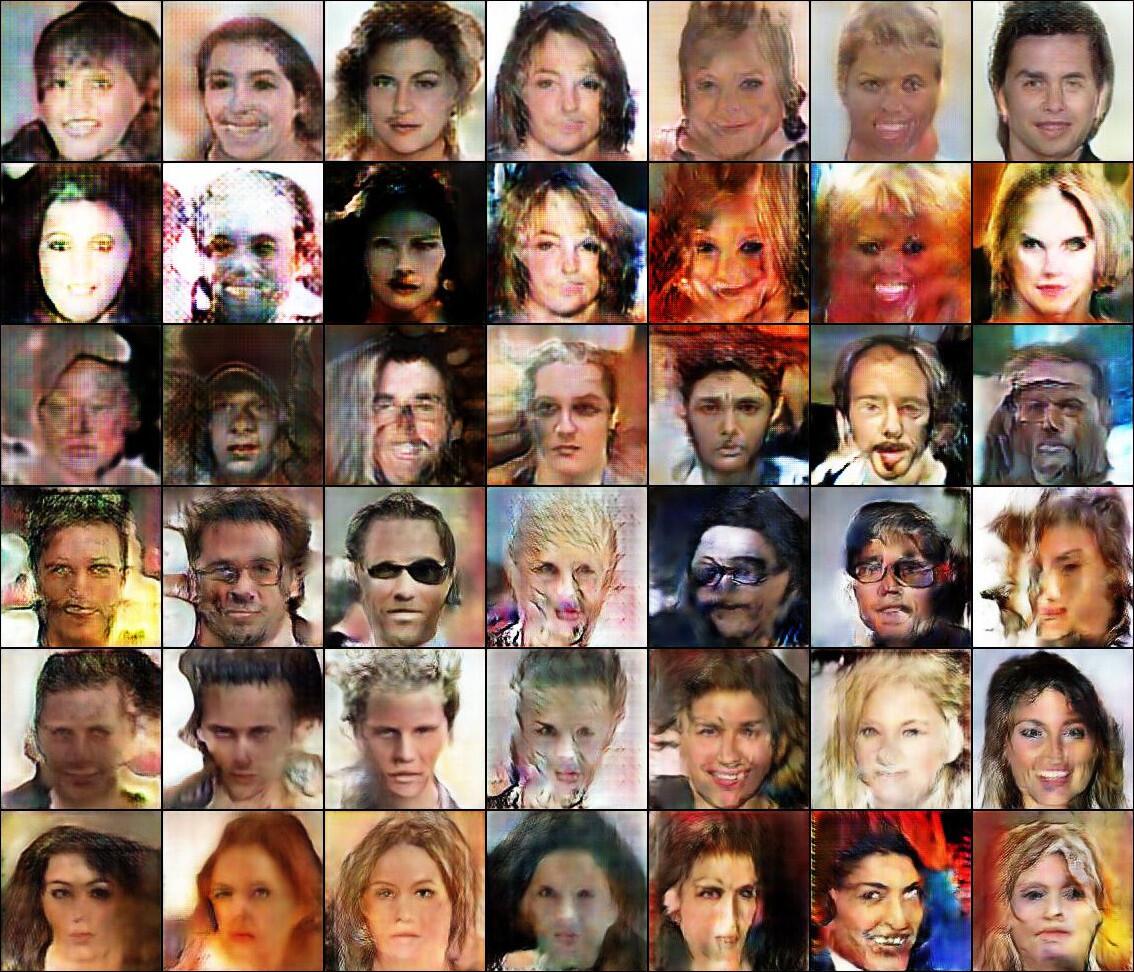}
\end{center}
\caption{\label{fig:resnet_stability} Evolution of samples during ResNet training. Top 3 rows: vanilla;  Middle 3 rows: BN; Bottom 3 rows: WN. Columns are samples from 10,000 to 70,000 iterations at intervals of 10,000 iterations. It is hardly recognizable but samples in the same row are indeed generated from the same code.}
\end{figure}

\begin{figure}
\begin{center}
\includegraphics[width=\linewidth]{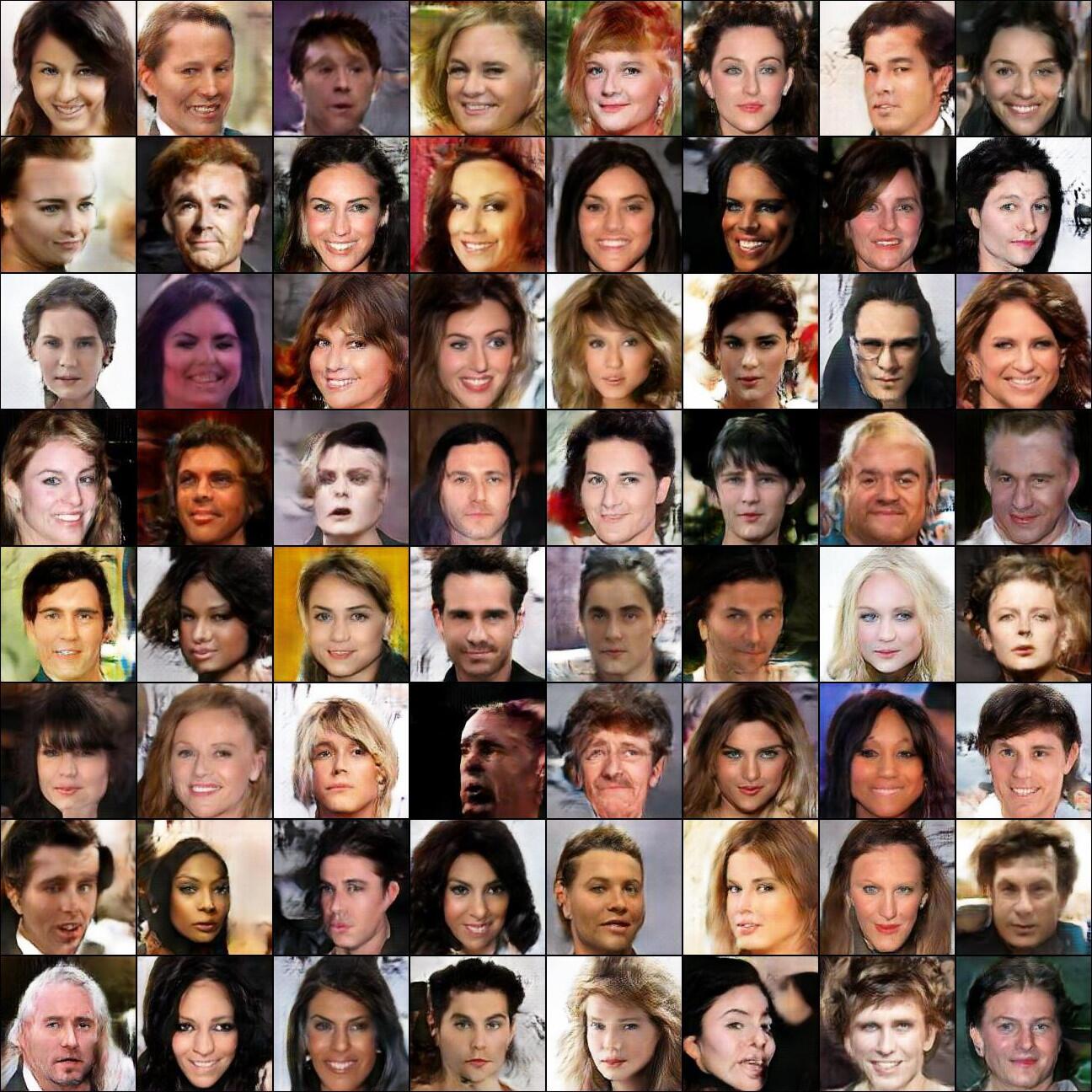}
\end{center}
\caption{\label{fig:resnet_samples_wn} Random samples from WN ResNet model, at iteration 195,000.}
\end{figure}

During 70,000 iterations of training, the vanilla model and the BN model were never able to generate more than a handful of different samples (random samples from iteration 70,000 shown in figure \ref{fig:resnet_samples}) and were extremely unstable (evolution of samples for every 10,000 iterations shown in figure \ref{fig:resnet_stability}).
The WN model was trained to iteration 300,000 without major issues, and was able to generate samples with high quality and diversity. The best running reconstruction loss was 0.016906, achieved at iteration 195,000. Random samples from that iteration are shown in figure \ref{fig:resnet_samples_wn}.

We do point out however, that with continued training after around iteration 200,000 we observe some degradation of sample quality in the weight normalized ResNet model, in similar ways as in the vanilla DCGAN model examined in Appendix \ref{sec:moresamples}. This indicates that Weight Normalization in itself may not be sufficient to guarantee the stability of the network. But it is not our goal to compete with other techniques and find a complete solution to the instability of GAN training. Rather, since our method does not propose different training loss (e.g. least squares in LSGAN \citep{mao2016least}) or protocol (e.g. batch discrimination) or favour a particular architecture (e.g. autoencoder-based, in EBGAN \citep{zhao2016energy}), our method is complementary to these existing GAN training improvement techniques, and can be combined with any of these to further improve the quality of GANs.

Our method is not compatible with methods that operate on the weights, notably weight clipping in Wasserstein GAN \citep{arjovsky2017wasserstein}. But for this particular case, weight normalization provides an alternative way to weight clipping to enforce Lipschitz continuity, as discussed in Appendix \ref{sec:wgan}.

\section{Conclusion}

We introduced weight normalization for the training of GANs using an alternative formulation than the original work of \citep{salimans2016weight} and \citep{arpit2016normalization}, which achieves superior training performance.
We also presented an evaluation method for GANs based on the mean squared Euclidean distance between the test samples and the closest generated ones, which are synthesized via gradient descent on a latent code.
We trained and analyzed variants of DCGAN \citep{radford2015unsupervised} with different normalization methods for image generation on datasets of multiple scales. We found that batch-normalized models perform worse in reconstructing test samples and are less stable during training. In particular, both reconstruction errors and the visual quality can be deteriorated by BN. However, our formulation of weight normalization improves both reconstruction quality and training stability considerably. We further demonstrate the stabilizing power of weight normalization by successful training of a residual GAN that is considerably deeper. Based on our extensive evaluations, we believe that weight normalization should be used instead of batch normalization when training generative adversarial networks.

\bibliography{weightnorm}
\bibliographystyle{ieee}

\appendix
\ifcvprfinal\else
\newpage
\fi
\section{Proof for The Equivalence Between a Non-Normalized and a Strict Weight-Normalized Network with One Affine Weight-Normalized Layer at the End}
\label{sec:equivalence}

Consider two networks with $(2n+1)$ layers each. The first network is a non-normalized network, where layers $(2k+1)\ (0\le k\le n)$ are linear layers and layers $2k\ (1\le k\le n)$ are ReLU layers. The second network is a weight-normalized network, where layers $(2k+1)\ (0\le k\le n-1)$ are strict weight-normalized layers, layer $(2n+1)$ is an affine weight-normalized layer, and layers $2k\ (1\le k\le n)$ are translated ReLU layers.

We make the following claim:

\newtheorem{equiv0}{Claim}
\begin{equiv0}
\label{thm:equiv0}
The aforementioned two networks are capable of representing the same set of functions.
\end{equiv0}

First, we prove that a linear-and-ReLU combination is equivalent to a strict weight-normalized-and-TReLU combination, if both are augmented by a learned affine transformation at the end.

\newtheorem{equiv1}[equiv0]{Lemma}
\begin{equiv1}
A linear layer, followed by a ReLU layer, followed by an affine transformation, is equivalent to a strict weight-normalized layer, followed by a TReLU layer, then by an affine transformation.
\end{equiv1}
\begin{proof}
For simplicity we consider the case where the first layer has only one output neuron. Then, a linear layer, followed by a ReLU layer, followed by an affine transformation, becomes
\begin{equation}
y=\mathrm{ReLU}\left(\mathbf{w}^T\mathbf{x}+\alpha\right)\cdot\gamma+\beta
\end{equation}
while a strict weight-normalized layer, followed by a TReLU layer, followed by an affine transformation would be
\begin{equation}
y=\mathrm{TReLU}_{\alpha'}\left(\frac{\mathbf{w'}^T\mathbf{x}}{||\mathbf{w'}||}\right)\cdot\gamma'+\beta'
\end{equation}
where $\mathbf{w}, \mathbf{w}', \alpha, \alpha', \beta, \beta', \gamma \textrm{ and } \gamma'$ are learned parameters. The transformation
\begin{equation}
\begin{cases}
\mathbf{w'}&=\mathbf{w}\\
\alpha'&=-\frac{\alpha}{||\mathbf{w}||}\\
\beta'&=\beta+\alpha\cdot\gamma\\
\gamma'&=||\mathbf{w}||\cdot\gamma
\end{cases}
\quad\textrm{and}\quad
\begin{cases}
\mathbf{w}&=\mathbf{w}'\\
\alpha&=-||\mathbf{w}'||\cdot\alpha'\\
\beta&=\beta'+\alpha'\cdot\gamma'\\
\gamma&=\frac{\gamma'}{||\mathbf{w}'||}
\end{cases}
\end{equation}
establishes a one-to-one correspondence between these two forms.
\end{proof}

We make the following observations:

\newtheorem{equiv2}[equiv0]{Lemma}
\begin{equiv2}
A linear layer preceded by an affine transformation is equivalent to a single linear layer.
\end{equiv2}

\newtheorem{equiv3}[equiv0]{Lemma}
\begin{equiv3}
A linear layer is equivalent to an affine weight-normalized layer.
\end{equiv3}

These proofs are trivial.
Now we demonstrate the proof of claim \ref{thm:equiv0} by transforming network 1 to network 2:

We perform the following procedure for each $k$ from $1$ to $(n-1)$: first, we add an affine transformation between the ReLU layer $2k$ and the linear layer $(2k+1)$. We then exchange the linear layer $(2k-1)$ and the ReLU layer $2k$ with a strict weight-normalized layer and a TReLU layer. The additional linear transformations are then removed. By adding and removing an affine transformation does not change the expressiveness of the network since a linear layer succeeds it. With an affine layer in place, the exchange would not change the expressiveness of the network either.

Finally, we change the last linear layer $(2n+1)$ to an affine weight-normalized layer.

\section{Implementation Details}
\label{sec:detail}
Here we provide some implementation details regarding the weight normalized layers.

Note that for strided and transposed convolutional layers, each element in the output tensor receives an input from only a subset of the $c_i \times k_w \times k_h$ elements in the input tensor, which corresponds to the kernel, where $c_i$ is the number of input features and $k_w$ and $k_h$ are the kernel width and height. 
Ideally, we should perform weight normalization for each of these different subsets of weights separately. In our experiments, we use a simple trick: we compute the norm of the weight for the full kernel as a whole, and divide the norm by $\sqrt{d_w \times d_h}$ where $d_w$ and $d_h$ are horizontal and vertical strides. This norm is used to normalize the weight in all different subsets.

The first layer in the generator deserves some special treatment. While it can be seen as a transposed convolutional layer, (since the spatial size of the input is $1 \times 1$), it can also be viewed as a fully connected layer (with shared bias between output elements from the same feature map). These two views do make a difference when Weight Normalization is in place: similar to the case above, each output element actually receives an input from a subset of $c_i$ elements instead of $c_i \times k_w \times k_h$ elements, corresponding to the kernel. Hence, it is more appropriate to implement this layer as a weight normalized fully connected layer, which is what we did in our experiments.

We also found that weight initialization of the first layer of the generator had an impact on the effect of WN. In our experiments, initial weights are drawn uniformly from $[-0.01/\sqrt{c_i}, 0.01/\sqrt{c_i}]$ where $c_i$ is the size of the input which is just the length of the latent code. For the convolutional layers, initial weights are drawn uniformly from $[-1/\sqrt{c_i\cdot k_w\cdot k_h}, 1/\sqrt{c_i\cdot k_w\cdot k_h}]$, as usual.

When computing the norm, a numerical stability term $\varepsilon=10^{-6}$ is added to the sum of squared weights before taking the square root.

Recall that the purpose of weight normalization is to normalize the mean and variance of the output of a linear layer. In the strict weight normalized case (equation \ref{eqn:strictweightnorm}), if each dimension of the input vector $\mathbf{x}$ is independently drawn from a distribution with expected value 0 and variance 1, then the output $y$ will also have expected value 0 and variance 1.

In a residue block, if the shortcut branch and the residue branch are simply summed, the output distribution will have variance 2, and the normalizing effect would be lost. One possible fix to this is to simply divide the sum by $\sqrt{2}$, but this will cause the shortcut branch to vanish as the network grows deeper, which defeats the purpose of residual networks. Our solution is what we call ``weight normalized addition''. Consider first the simplest case of adding two variables. We take
\begin{equation}
y = \frac{w_1 x_1 + w_2 x_2}{\sqrt{w_1^2 + w_2^2}}
\end{equation}
where $w_1$ and $w_2$ are learned weights. This will preserve the normalizing effect. At the same time, if we set the initial weight to 0 on the residue branch and 1 on the shortcut branch, the shortcut branch will be able to ``go through'' the network at the beginning, thus the benefits of residual networks are also preserved.

When adding two convolutional feature maps, we learn pairs of weights for each feature channel and share the weights across spatial locations.

\section{Additional Samples and Analysis}
\label{sec:moresamples}

Here we show addtitional generated samples from the three models. In addition, we talk about certain issues with GAN training that can only be detected by visually inspecting large amounts of samples. 

\subsection{Vanilla Model}

\begin{figure}
\begin{center}
\includegraphics[width=\linewidth]{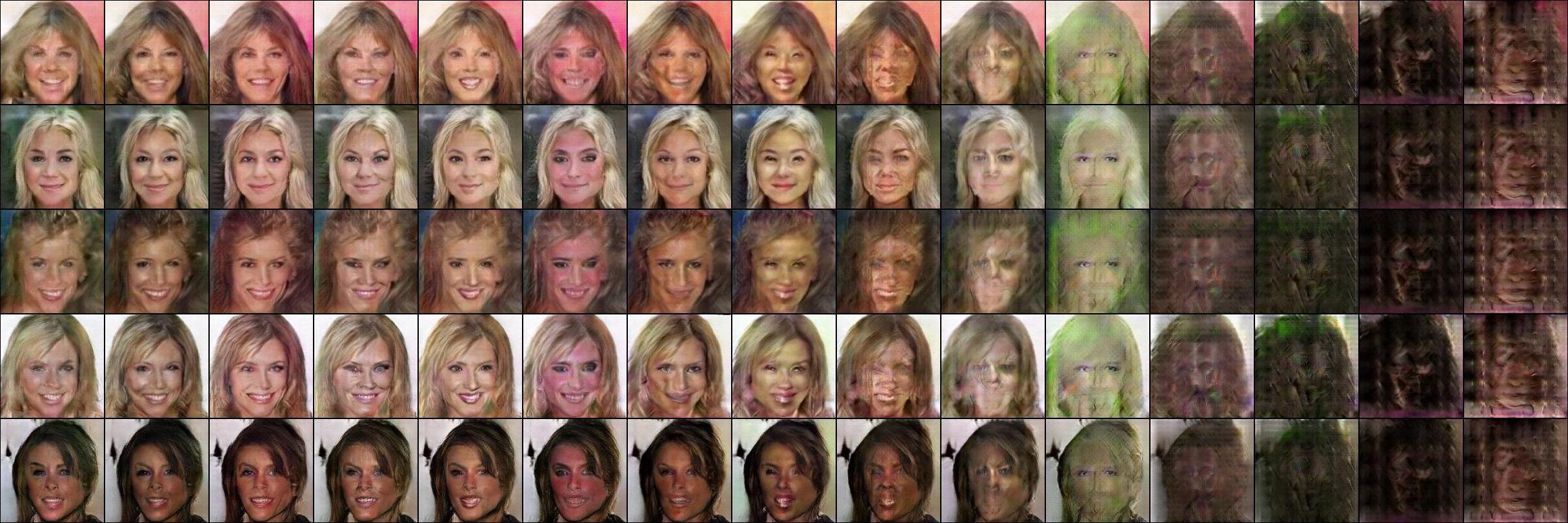}
\end{center}
\caption{\label{fig:nonorm_run1_collapse} Samples generated by vanilla model, every 100 iterations from 134,000 to 135,400}
\end{figure}

\begin{figure}
\begin{center}
\includegraphics[width=\linewidth]{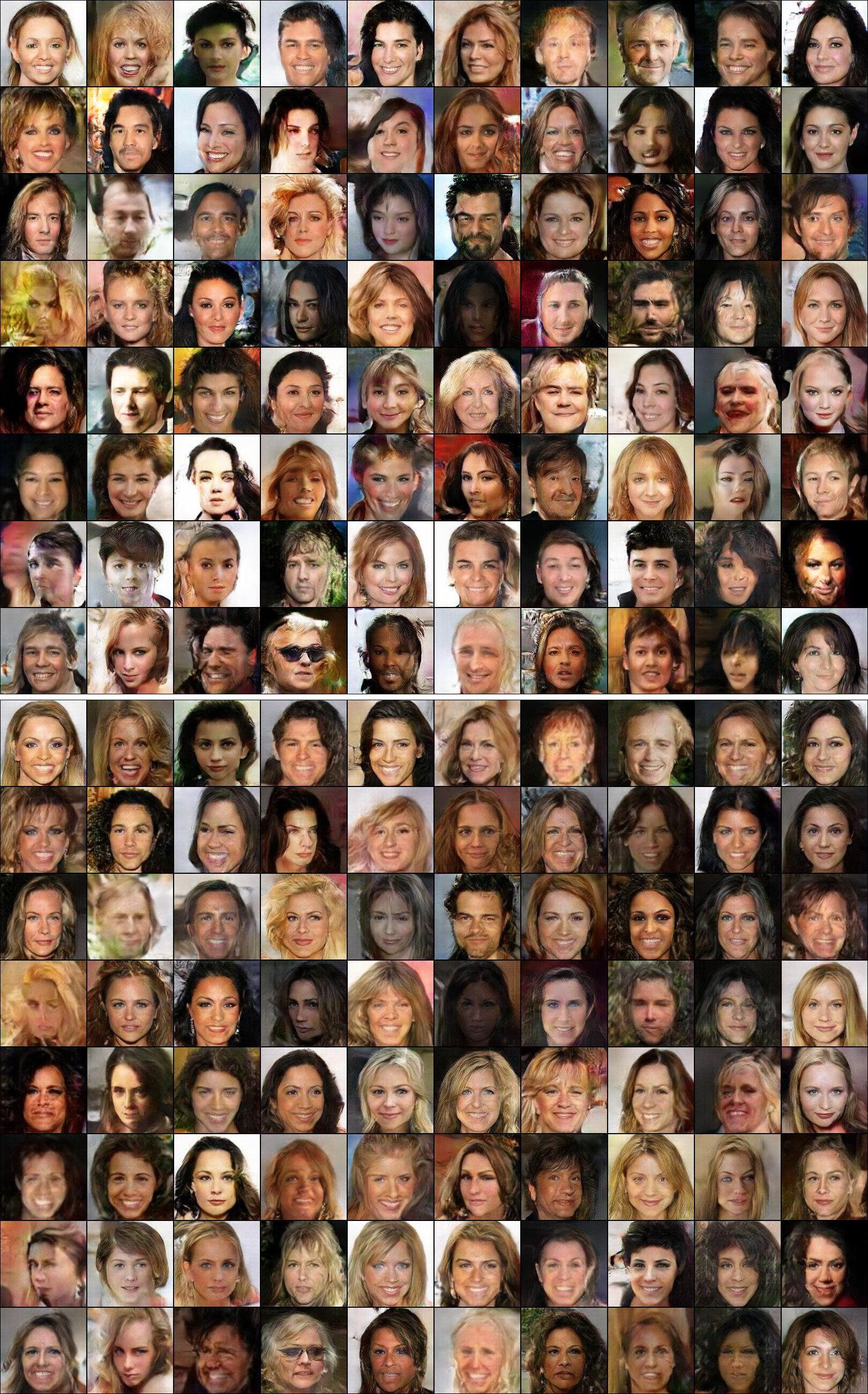}
\end{center}
\caption{\label{fig:nonorm_run1} Random samples generated by vanilla model. Upper half: iteration 30,500. Lower half: Iteration 120,000.}
\end{figure}

Figure \ref{fig:nonorm_run1} shows samples generated by the vanilla model from the same set of random codes, at iteration 30,500 (optimal iteration) and 120,000. Samples generated at 120,000 iterations are visually superior on average if each sample is inspected individually, despite a higher reconstruction loss. But notice how the diversity of the samples has decreased: the lower half of the figure is dominated by yellow and brown colors and are darker than the upper half. Even more subtle, the lower half has less variation in facial expressions. The gender diversity is also decreasing: some clear male faces in the upper half become more feminine in the lower half.

When training beyond a certain amount of iterations, the samples start to evolve toward the same direction. While the samples are still different, similar changes can be observed in each iteration. In this process, the difference between samples is gradually lost. This corresponds to the slow and steady increase of the reconstruction loss. At around 130,000 iterations, this process suddenly accelerates and then the model collapses at a certain point, as shown in figure \ref{fig:nonorm_run1_collapse}.

Interestingly, this appears to be like a reversed behavior of an early stage training. In particular, training usually starts with a code that generates a similar output and changing in a similar way until the synthesized samples start to gain diversity.

\begin{figure}
\begin{center}
\includegraphics[width=\linewidth]{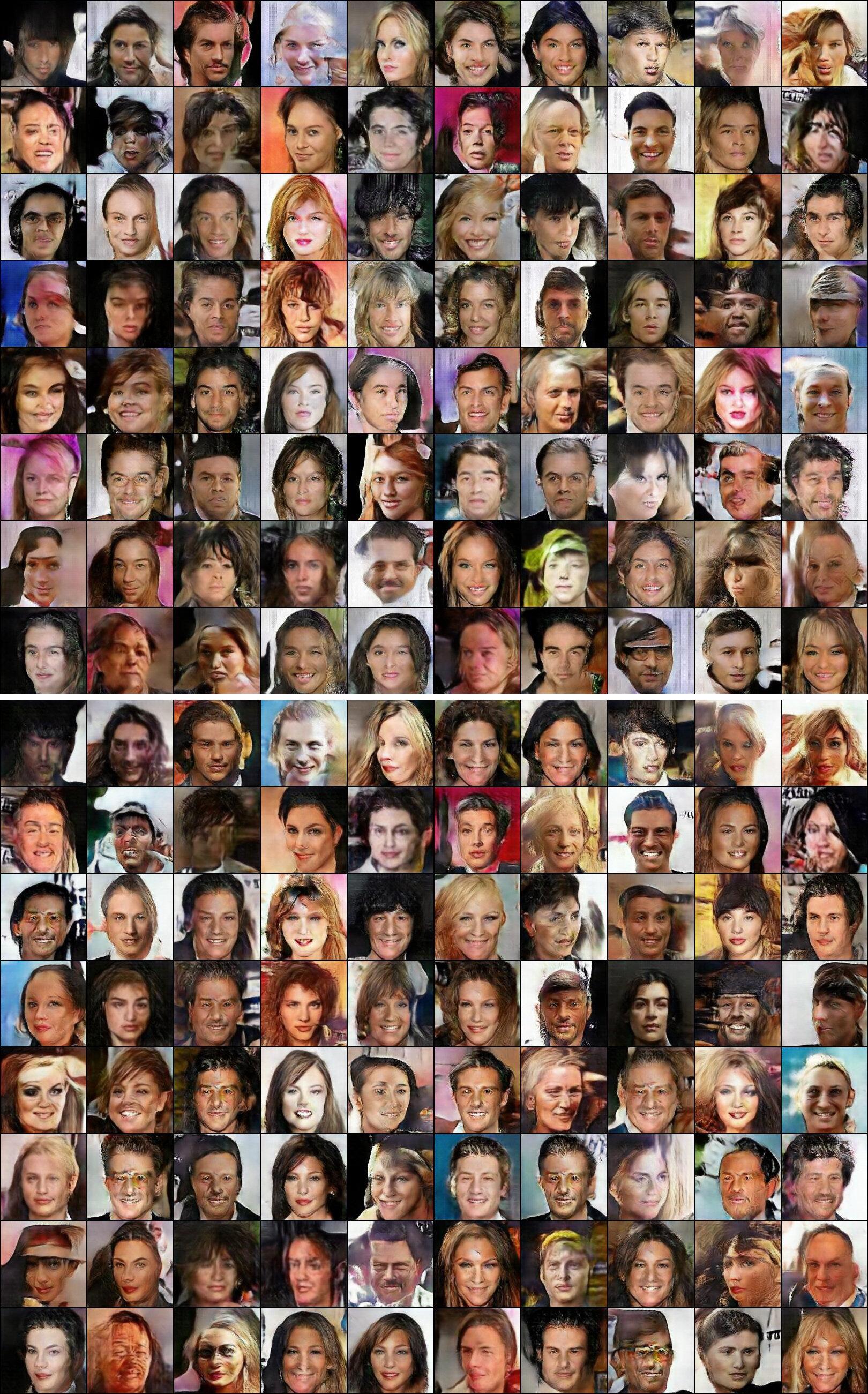}
\end{center}
\caption{\label{fig:batchnorm_run2} Random samples generated by BN model. Upper half: iteration 30,500. Lower half: Iteration 110,000.}
\end{figure}

\subsection{Batch-Normalized Model}

\begin{figure}
\begin{center}
\includegraphics[width=\linewidth]{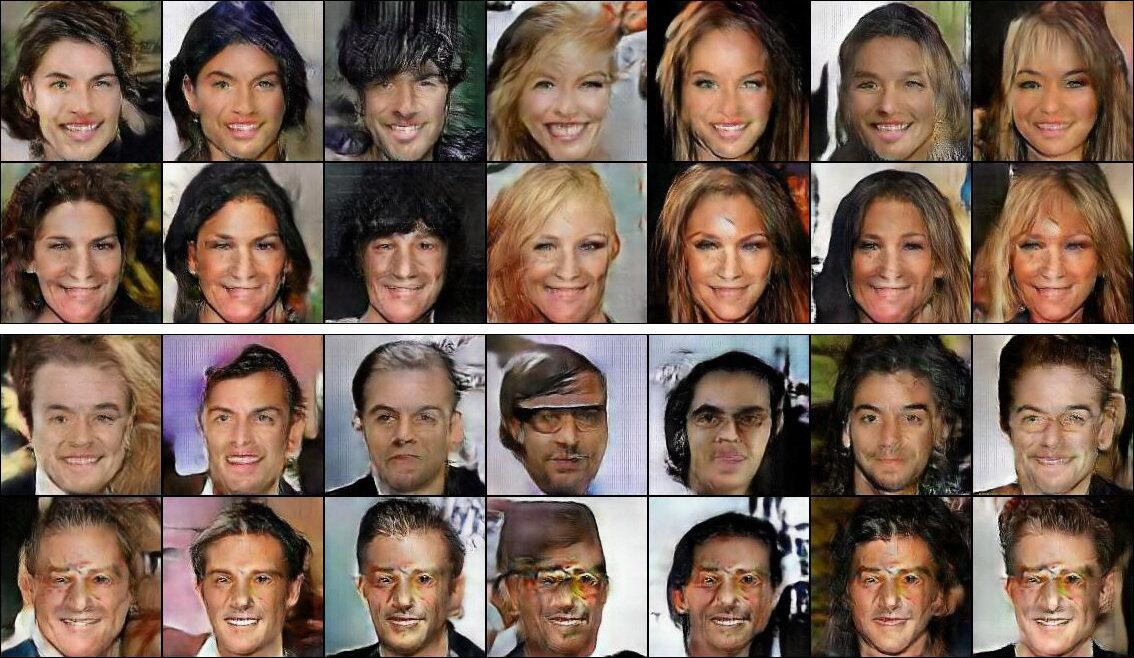}
\end{center}
\caption{\label{fig:batchnorm_run2_collapse} Selected samples from figure \ref{fig:batchnorm_run2}. Row 1 and 3 are from iteration 30,500. Row 2 and 4 are corresponding samples from iteration 110,000.}
\end{figure}

Figure \ref{fig:batchnorm_run2} shows samples generated by the BN model from the same set of random codes, at iteration 30,500 (optimal iteration) and 110,000. At first sight, it does not show a decrease in diversity as with the vanilla model. But after comparing the samples carefully, we can discover certain repeatedly-occurring features. To see this more clearly, in Figure \ref{fig:batchnorm_run2_collapse} we picked and rearranged several samples from Figure \ref{fig:batchnorm_run2}.

We identified two groups from samples generated at iteration 110,000. Within each group, while the appearance of the face varies considerably, almost the exact same expression is produced. When comparing these samples from ones that are generated from the same code at iteration 30,500, we found that a second group is new, while the first group has already existed for a long period of time. This indicates that the BN model had limited diversity even at its optimal iteration.

This is a indication of a different cause for mode collapse: as the training progresses, certain features become dominant. While most samples stay different, more and more start to acquire these dominating features. In the extreme case, only a handful of different possible outputs remain in the end.

\begin{figure}
\begin{center}
\includegraphics[width=\linewidth]{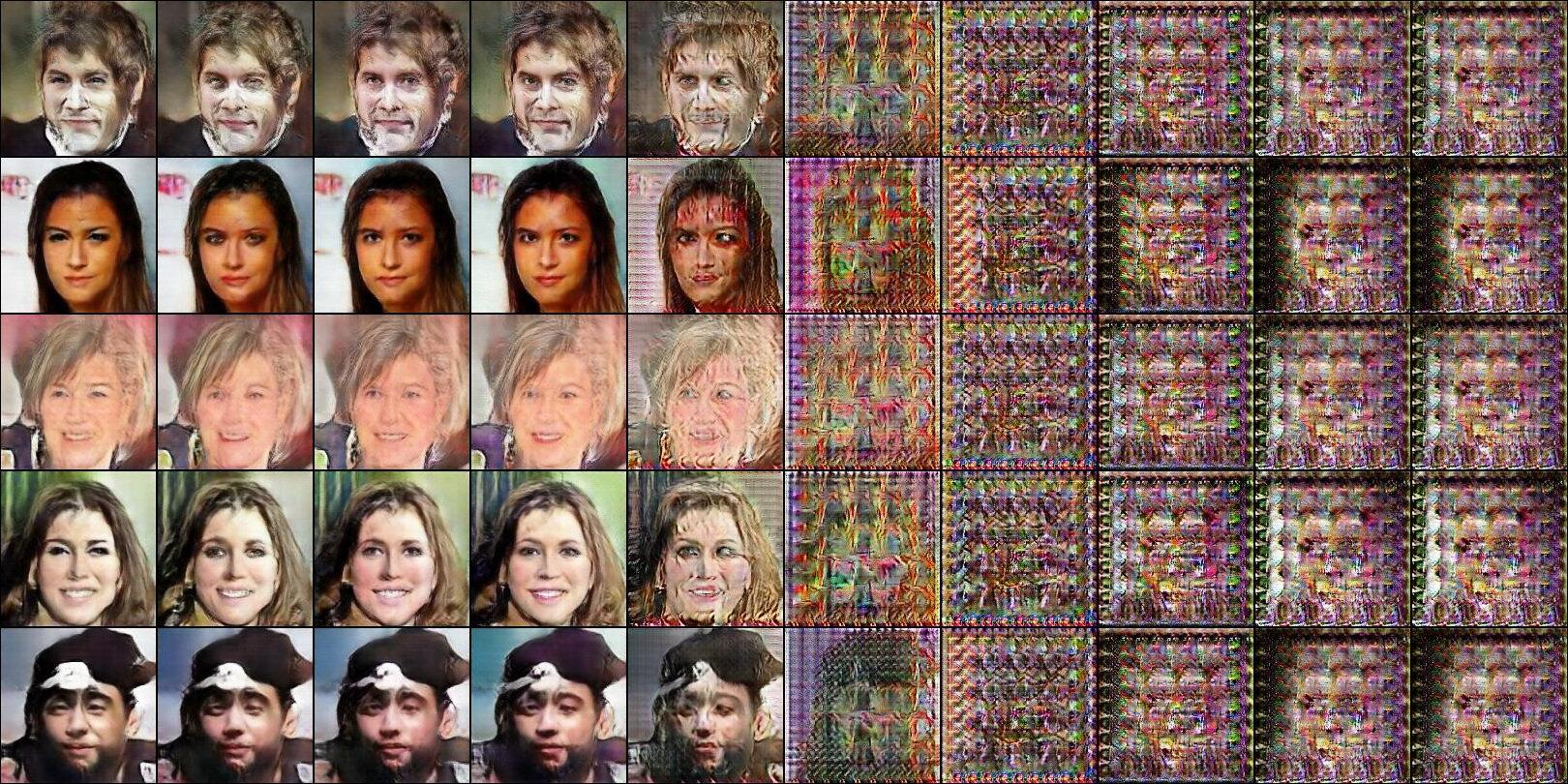}
\end{center}
\caption{\label{fig:batchnorm_run3_collapse} Samples generated by BN model instance 3, every 100 iterations from 38,300 to 39,200}
\end{figure}

Alternatively, mode collapse can suddenly happen, as is the case for the ``failed'' training instance 3 in Appendix \ref{sec:moreinstances}. We show the samples that are generated when the model collapses in Figure~\ref{fig:batchnorm_run3_collapse}. Notice that there was no clear sign of decreased diversity prior to the sudden collapse.

\subsection{Weight-Normalized Model}

\begin{figure}
\begin{center}
\includegraphics[width=\linewidth]{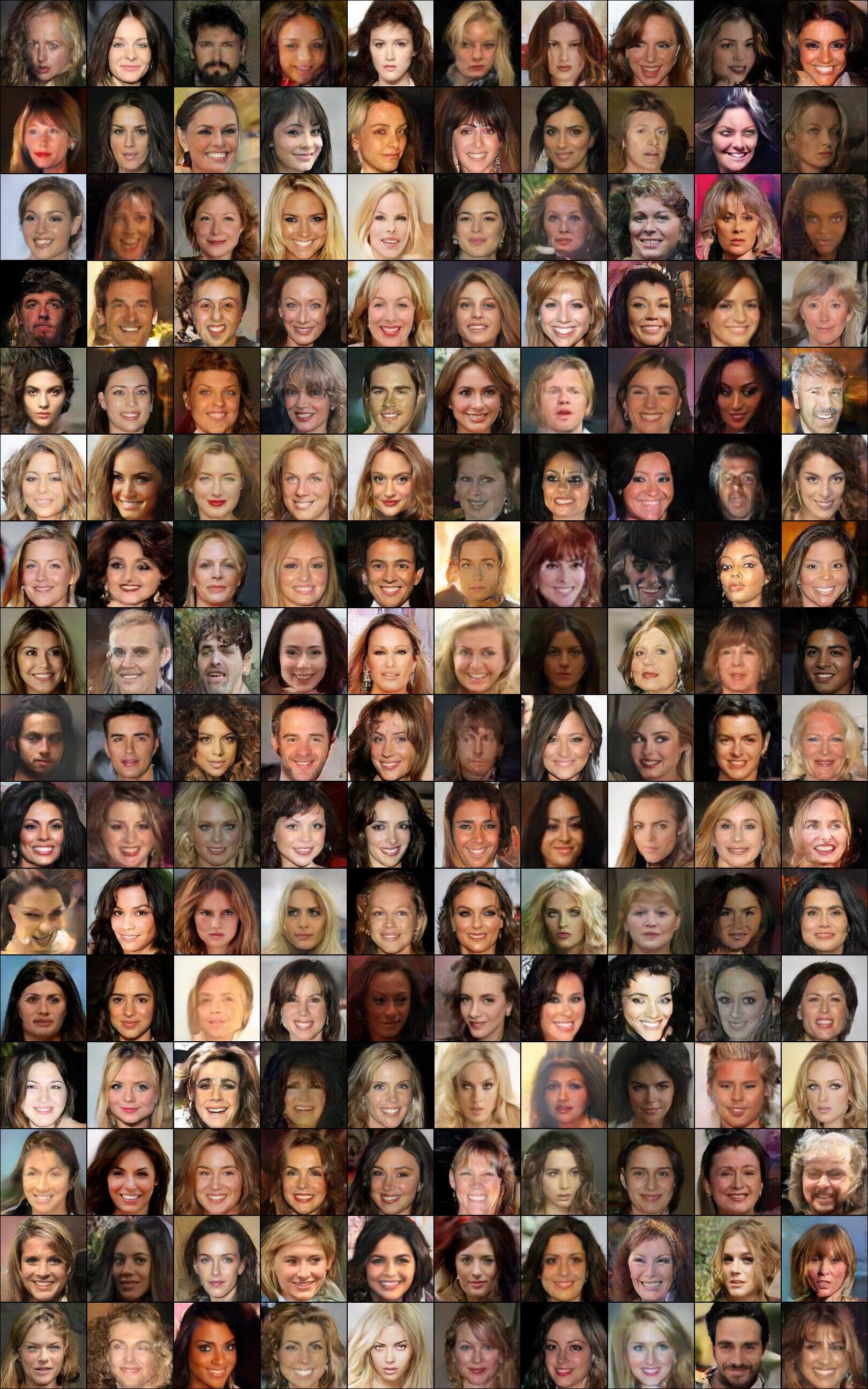}
\end{center}
\caption{\label{fig:weightnorm_run1} Random samples generated by WN model at iteration 463,000}
\end{figure}

Figure \ref{fig:weightnorm_run1} illustrates random samples generated by the optimal WN model. In terms of diversity, it is not ideal, as the samples still show a lack of color variation and an unbalanced gender ratio compared to the ground truth distribution.
However, they show more variations than the vanilla model and less subtle recurring features compared to the BN model. The individual samples are of higher quality on average as well. Also note the relative low rate of ``failed'' or highly implausible samples.

\begin{figure}
\begin{center}
\includegraphics[width=\linewidth]{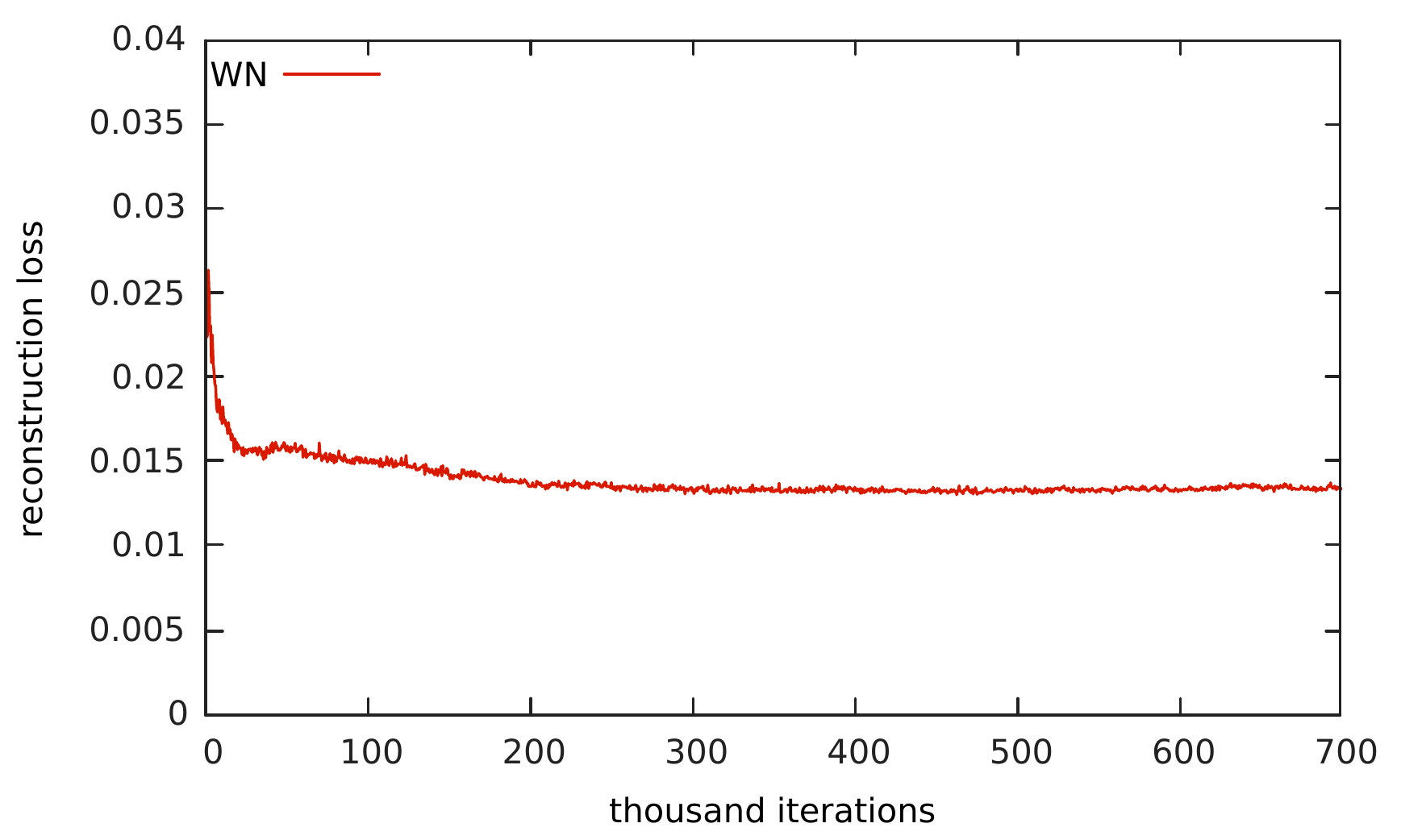}
\end{center}
\caption{\label{fig:reclosswn} Running reconstruction loss during training of WN model}
\end{figure}

Figure \ref{fig:reclosswn} shows the running reconstruction loss recorded during the whole training process of the WN model. The loss remains nearly constant after 300,000 iterations, which demonstrates the stability of the WN model.

\subsection{Random Reconstructed Samples}

\begin{figure}
\begin{center}
\includegraphics[width=\linewidth]{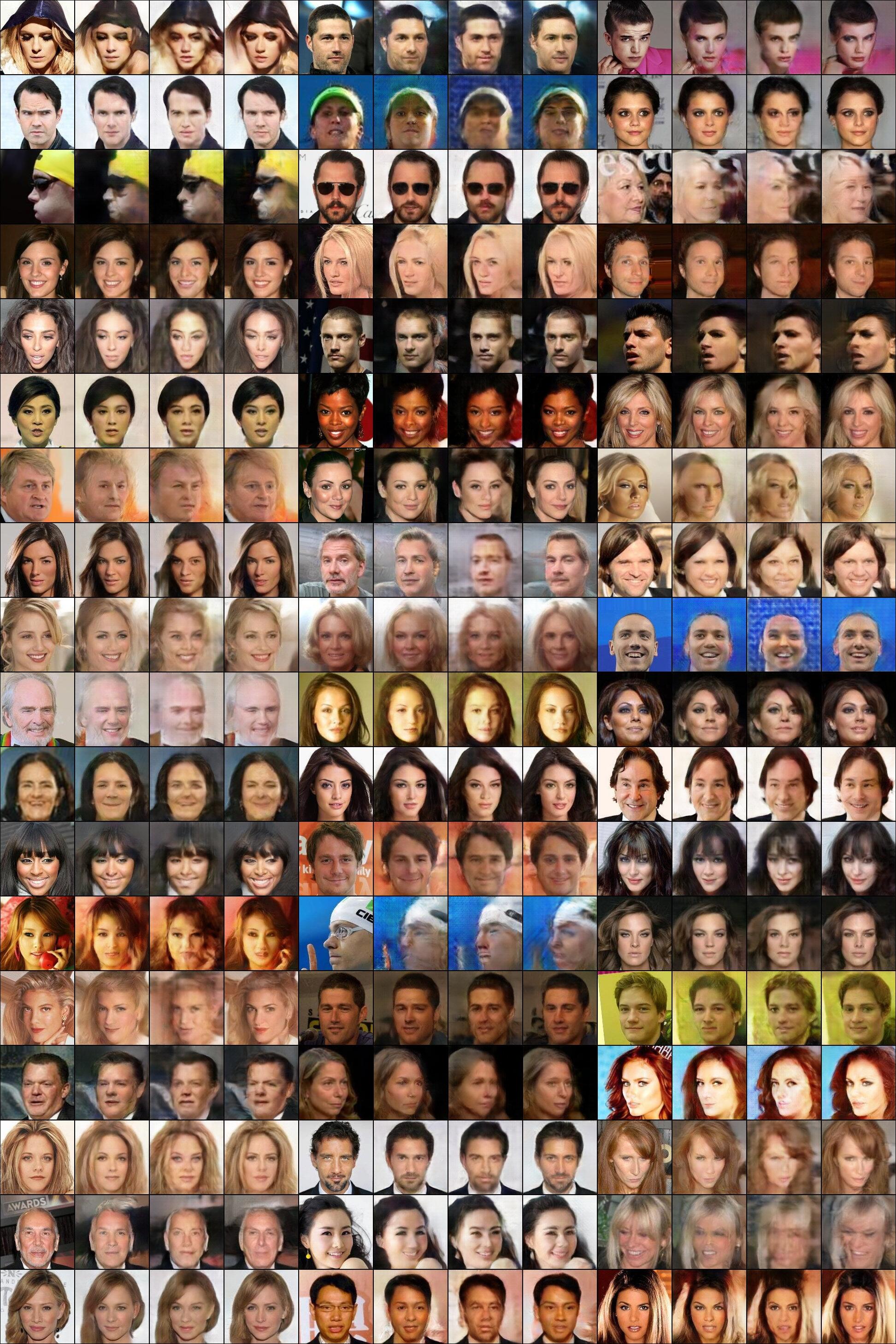}
\end{center}
\caption{\label{fig:rec_random} Randomly reconstructed test samples. From left to right in each group: test sample, vanilla reconstruction, BN reconstruction, WN reconstruction.}
\end{figure}

Figure~\ref{fig:rec_random} shows a random selection of reconstructed samples.

\section{Additional Experiments}
\label{sec:moreexperiments}
\subsection{Additional Training Instances of Vanilla and BN Models}
\label{sec:moreinstances}

\begin{figure}
\begin{center}
\includegraphics[width=\linewidth]{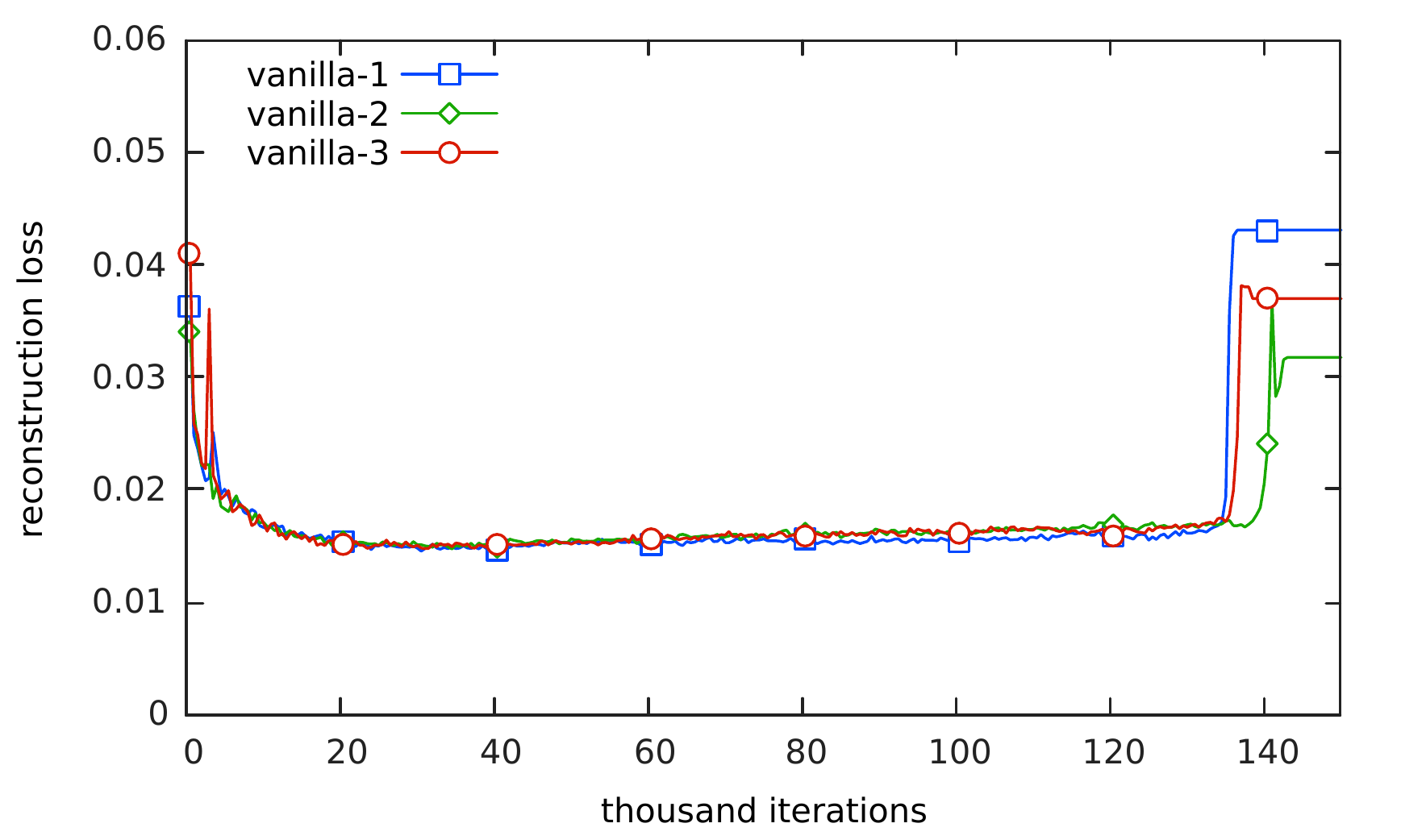}
\end{center}
\caption{\label{fig:recnonorm} Running reconstruction loss of repeated training of vanilla model.}
\end{figure}

\begin{figure}
\begin{center}
\includegraphics[width=\linewidth]{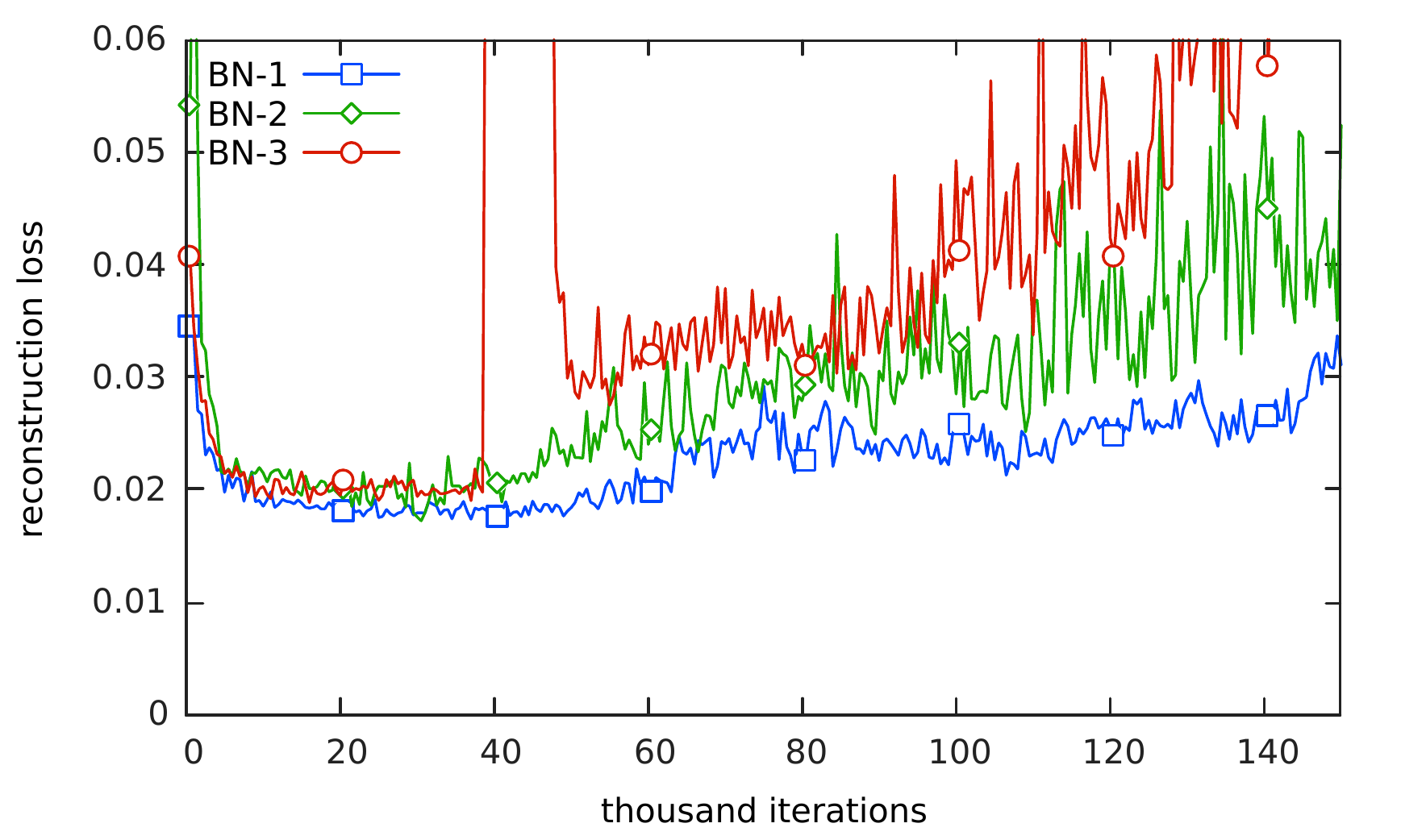}
\end{center}
\caption{\label{fig:recbatchnorm} Running reconstruction loss of repeated training of batch-normalized model.}
\end{figure}

Figures \ref{fig:recnonorm} and \ref{fig:recbatchnorm}, and Tables \ref{tab:recnonorm} and \ref{tab:recbatchnorm} show the reconstruction loss recorded during all training instances of the vanilla and batch-normalized models.

\begin{table}
\caption{\label{tab:recnonorm} Optimal reconstruction loss of repeated training of vanilla model.}
\begin{center}
\begin{tabular}{ccc}\hline
Instance  & Optimal iteration & Running loss \\\hline
vanilla-1 & 30,500            & 0.014509     \\
vanilla-2 & 34,500            & 0.014703     \\
vanilla-3 & 31,000            & 0.014734     \\\hline
\end{tabular}
\end{center}
\end{table}

\begin{table}
\caption{\label{tab:recbatchnorm} Optimal reconstruction loss of repeated training of batch-normalized model.}
\begin{center}
\begin{tabular}{ccc}\hline
Instance & Optimal iteration & Running loss \\\hline
BN-1     & 37,000            & 0.017349     \\
BN-2     & 30,500            & 0.017199     \\
BN-3     & 16,000            & 0.018810     \\\hline
\end{tabular}
\end{center}
\end{table}

We can see that the three instances of vanilla model gave almost identical loss curves. They achieved similar optimal loss at similar times, and the mode collapse also happened around the same time.

In addition to the instability observed for each training instance, the batch-normalized models also showed ``meta-instability'' as the behavior differed considerably between each training instance. Notably, in the third instance, mode collapse happened very early on. The training did recover to some extent, but the model was never able to regain the same sampling diversity as before the mode collapse.

\subsection{Additional Models}
\label{sec:moremodels}

For completeness, we also compare different formulations of Weight Normalization. The first one is a full-affine WN model, constructed from the WN model by replacing all strict weight-normalized layers by affine weight-normalized layers and all TPReLU layers by PReLU layers. The second one is a model with Weight Normalization plus mean-only Batch Normalization, as used in \citep{salimans2016weight}, constructed by taking the affine WN model and adding mean-only Batch Normalization at those places where regular Batch Normalization layers would be used in the BN model.

\begin{figure}
\begin{center}
\includegraphics[width=\linewidth]{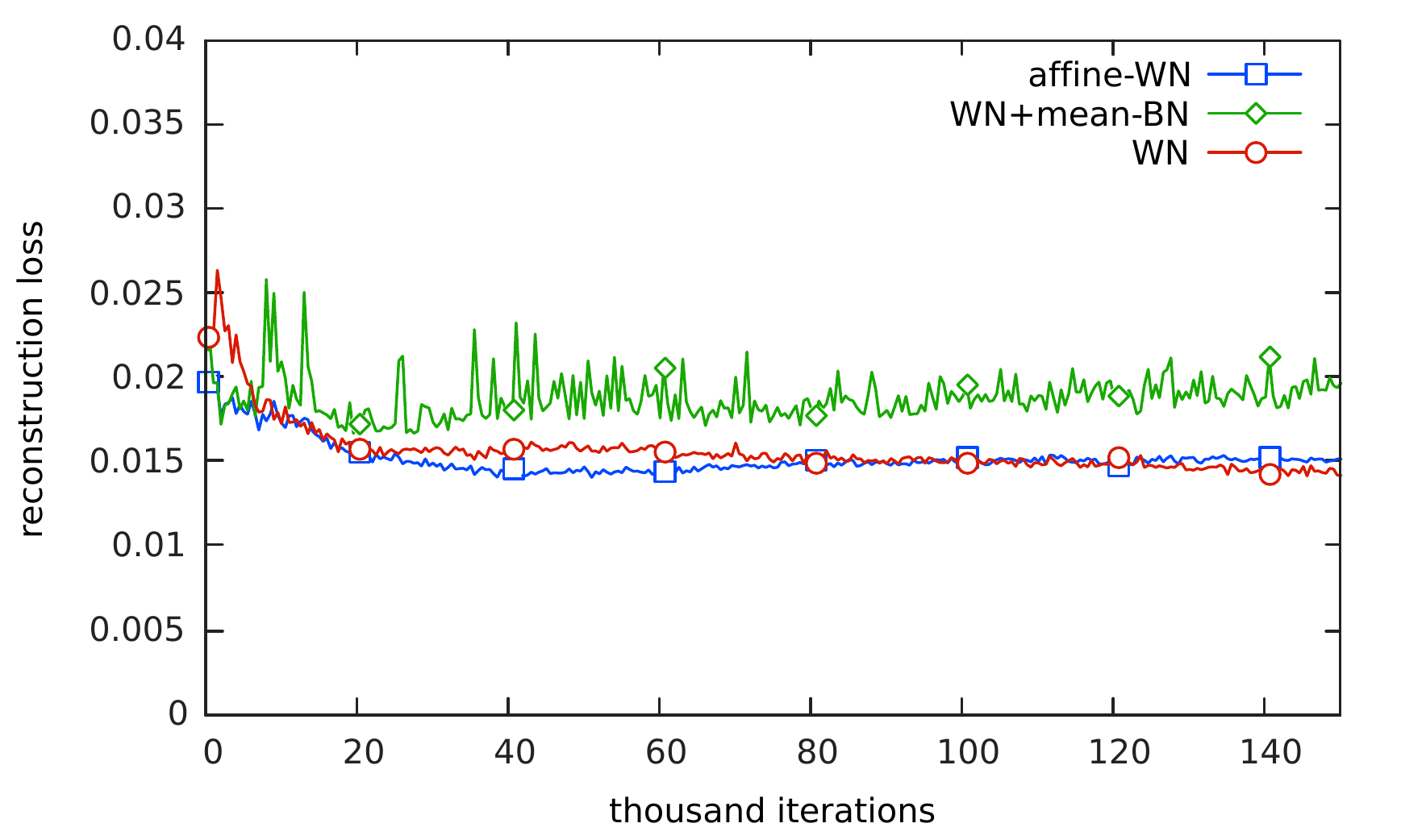}
\end{center}
\caption{\label{fig:recextra} Running reconstruction loss during training of additional models.}
\end{figure}

The reconstruction loss for the first 150,000 steps are shown in Figure~\ref{fig:recextra} and Table~\ref{tab:recextra}. The results of the WN model are included for comparison.

\begin{table}
\caption{\label{tab:recextra} Optimal reconstruction loss of additional models.}
\begin{center}
\begin{tabular}{ccc}\hline
Model       & Optimal iteration & Running loss \\\hline
affine-WN   & 51,000            & 0.014034     \\
WN+mean-BN  & 19,500            & 0.016639     \\
WN          & 463,000           & 0.013010     \\\hline
\end{tabular}
\end{center}
\end{table}

The WN with mean-only BN model achieved worse reconstruction than the vanilla model. In addition, although less severe than the BN model, it results in similar fluctuations of the BN model. We believe that the major advantage of WN over BN is its independence from batch statistics. By adding mean-only BN, this dependency is re-introduced, which harms the stability of the model.

\begin{figure}
\begin{center}
\includegraphics[width=\linewidth]{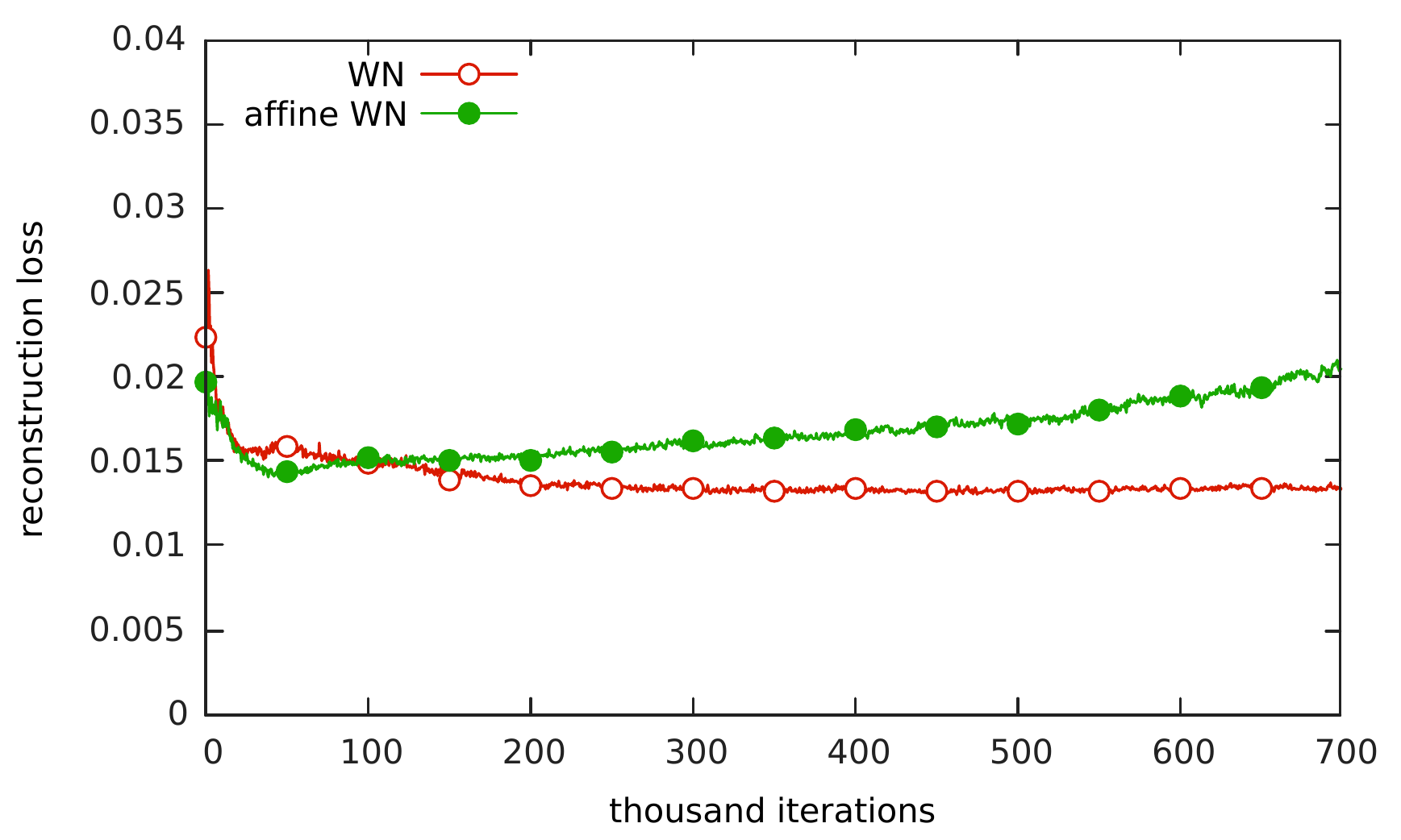}
\end{center}
\caption{\label{fig:reclosswnvsawn} Running reconstruction loss during training of the WN and affine-WN models}
\end{figure}

We are also interested in the comparison between the WN model and the affine-WN model, as it compares our formulation of Weight Normalization against the one originally one in \citep{salimans2016weight}. For this purpose, we trained the affine-WN model using also 700,000 iterations. Figure \ref{fig:reclosswnvsawn} shows the comparison of their reconstruction loss during training.

After making a quick descent in the beginning, the running reconstruction loss of the affine-WN model start to increase steadily. Unlike the vanilla model, the generated samples kept stable and are of high-quality. Hence, we evaluate both the optimal model and the model at iteration 700,000. The results are compared with the WN model in table \ref{tab:reclosswnvsawn}.

\begin{table}
\caption{\label{tab:reclosswnvsawn} Optimal and converged reconstruction loss of WN and affine-WN models}
\begin{center}
\begin{tabular}{cccc}\hline
Model     & Iteration & Running loss & Final loss     \\\hline
affine-WN & 51,000    & 0.014034     & 0.005941       \\
affine-WN & 700,000   & 0.020478     & 0.005939       \\
WN        & 463,000   & 0.013010     & 0.005525       \\\hline
\end{tabular}
\end{center}
\end{table}

Surprisingly, the affine-WN model at iteration 700,000 yields equally good 2,000-step reconstructions as with iteration 51,000, when the model achieved optimal 50-step reconstruction. Both of them, however, are about 7.5\% worse than the WN model.

\subsection{Experiments on CIFAR-10 Dataset}
\label{sec:cifar}

There are 60,000 images (training plus validation) of size 32$\times$32 in the CIFAR-10 dataset. We construct models in similar ways as for CelebA, but begin with 96 output channels for the first convolutional layer and stop further convolutions when the spatial size of the feature map reaches 4$\times$4. The length of the code (256) and training batch size (32) remains the same.

We use 58,000 images for training and 2,000 images for evaluation. During training, evaluation is performed every 1,000 training iterations on 400 images, with 50 gradient descent steps. Final evaluation is performed on the whole test set with 2,000 gradient descent steps.

\begin{figure}
\begin{center}
\includegraphics[width=\linewidth]{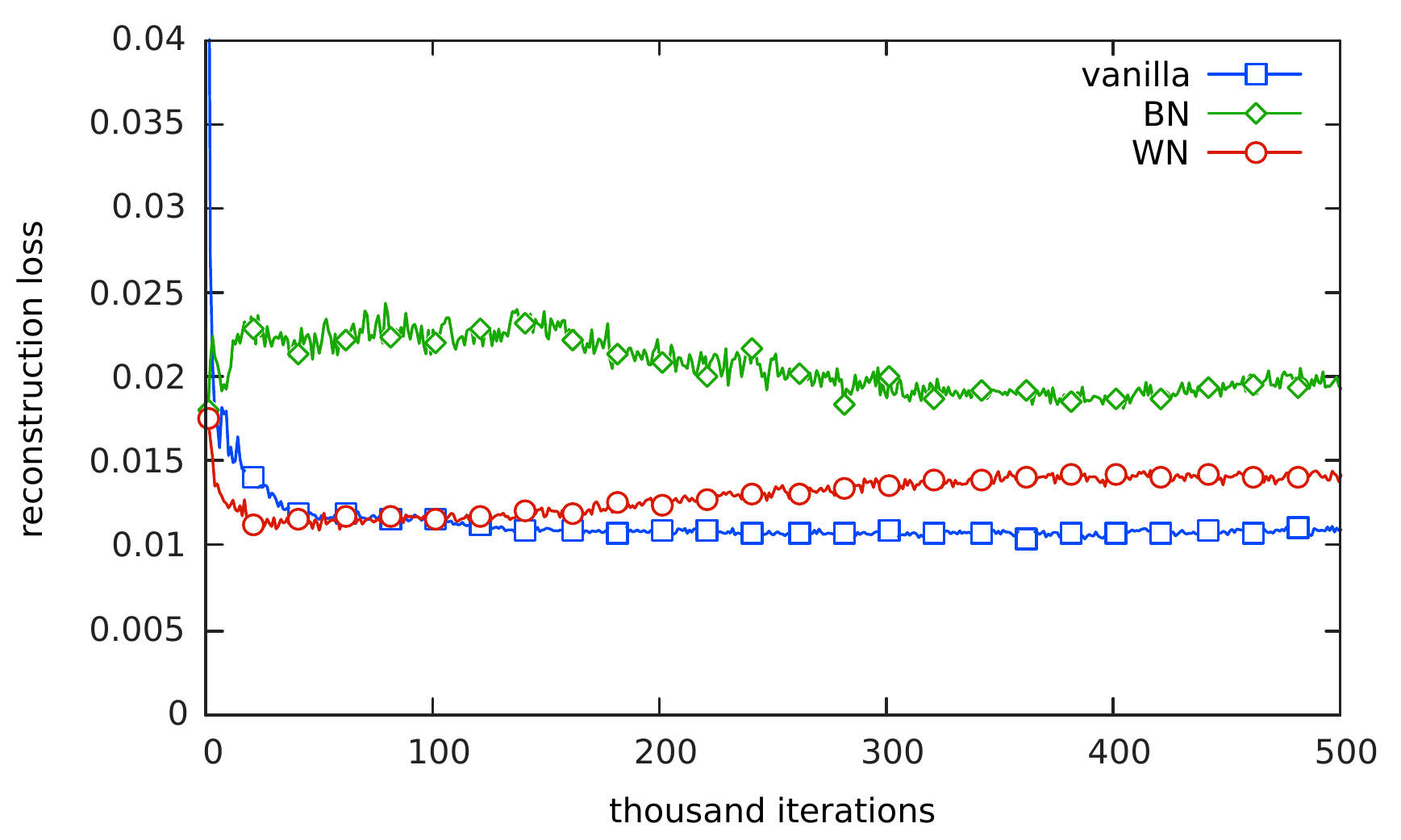}
\end{center}
\caption{\label{fig:cifarrecloss} Reconstruction loss during training on CIFAR-10.}
\end{figure}

BN is still the worst model compared to the vanilla and WN models. Now the WN model achieves optimal loss early on, but then becomes worse. On the other hand, the vanilla model keeps improving. However, both models converges, as shown by the flat section in the loss curve, between iterations 400,000 and 500,000. So we take these two models at iteration 500,000 for evaluation in additional to the optimal 50-step models. The BN model does not actually converge, as the rapidly changing ``recurring feature'', discussed in Section~\ref{sec:moresamples}, occurs. For completeness however, we also take the BN model from iteration 500,000 for evaluation.

The seemingly worse 500,000-iteration WN model turned out to give the best final reconstruction result. A more careful examination of the reconstruction process revealed that the 500,000-iteration WN model achieved better reconstruction than the optimal vanilla model at around 400th reconstruction step. We acknowledge that this exposes a weakness of our evaluation method: performing the reconstruction for too few steps may give inaccurate results, while too many steps would be time-consuming, which makes it unsuitable for training process monitoring.

\begin{table}
\caption{\label{tab:cifarrecloss} Optimal and converged reconstruction loss of the models on CIFAR-10}
\begin{center}
\begin{tabular}{cccc}\hline
Model   & Iteration & Running loss & Final loss \\\hline
vanilla & 387,000   & 0.010382     & 0.003413        \\
BN      & 1,000     & 0.017987     & 0.004904        \\
WN      & 50,000    & 0.010906     & 0.003509        \\
vanilla & 500,000   & 0.010948     & 0.003414        \\
BN      & 500,000   & 0.019287     & 0.005421        \\
WN      & 500,000   & 0.014195     & 0.003269        \\\hline
\end{tabular}
\end{center}
\end{table}

\begin{figure}
\begin{center}
\includegraphics[width=\linewidth]{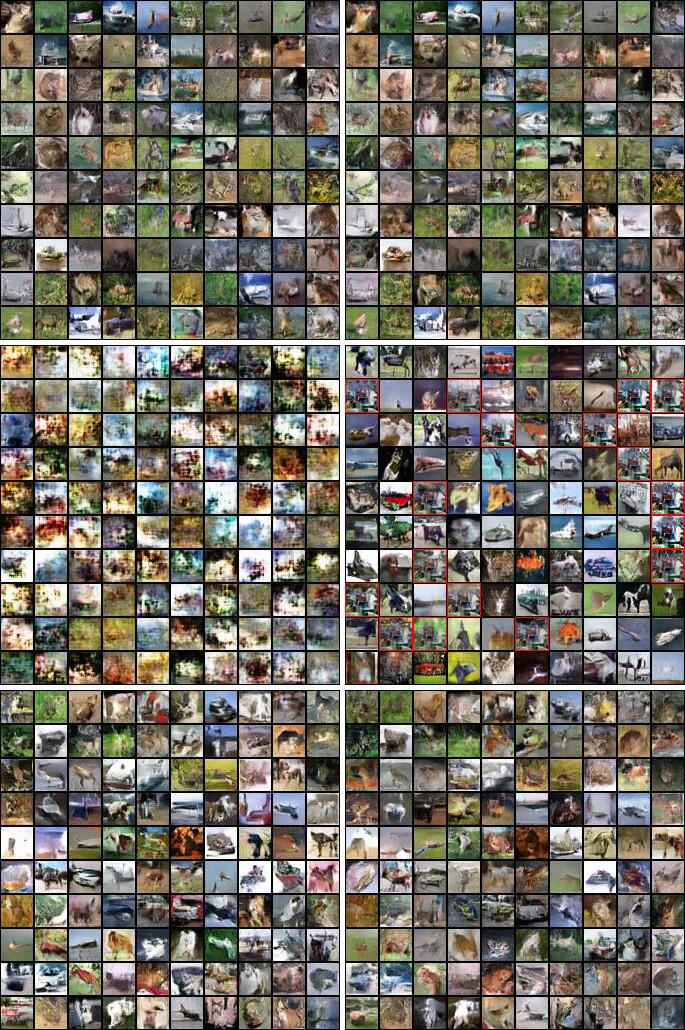}
\end{center}
\caption{\label{fig:cifarsamples} CIFAR-10 samples. Top to bottom: vanilla model, BN model, WN model. Left: optimal iteration. Right: iteration 500,000.}
\end{figure}

\begin{figure}
\begin{center}
\includegraphics[width=\linewidth]{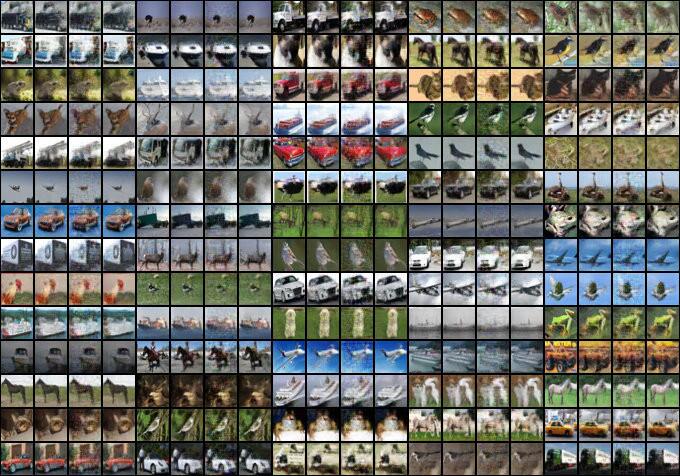}
\end{center}
\caption{\label{fig:cifarreconstructions} CIFAR-10 reconstructions. Each group, left to right: test sample, vanilla model, BN model, WN model.}
\end{figure}

Figure \ref{fig:cifarsamples} shows random samples generated by the three models, at their optimal iteration and at iteration 500,000. While the visual quality of samples from all models are good, the results are consistent in terms of diversity with the analysis in Appendix \ref{sec:moresamples}. The vanilla samples look dull and are dominated by one color (green); the BN samples show a recurring feature (marked with red border).

Figure \ref{fig:cifarreconstructions} shows random test samples and reconstructed ones.

\subsection{Experiments on LSUN Bedroom Dataset}
\label{sec:lsun}

There are 3,033,042 images in the bedroom class of the LSUN dataset, with images having 256 pixels on the shorter side. Unlike many published results on this dataset, we use the full-sized images. We crop with centered 256$\times$256 patches but do not down-sample the image. We construct models in a similar way as with CelebA, but stop further convolution when the spatial size of the feature map reaches 4$\times$4 and use a code length of 512. Due to the large size of the images and the network, we reduce the batch sizes to 12 to save computation.

We use 2,000 images for evaluation and the rest for training. During training, evaluation is performed every 1,000 iterations on 200 images, with 50 gradient descent steps. Final evaluation is performed on the whole test set with 2,000 gradient descent steps.

\begin{figure}
\begin{center}
\includegraphics[width=\linewidth]{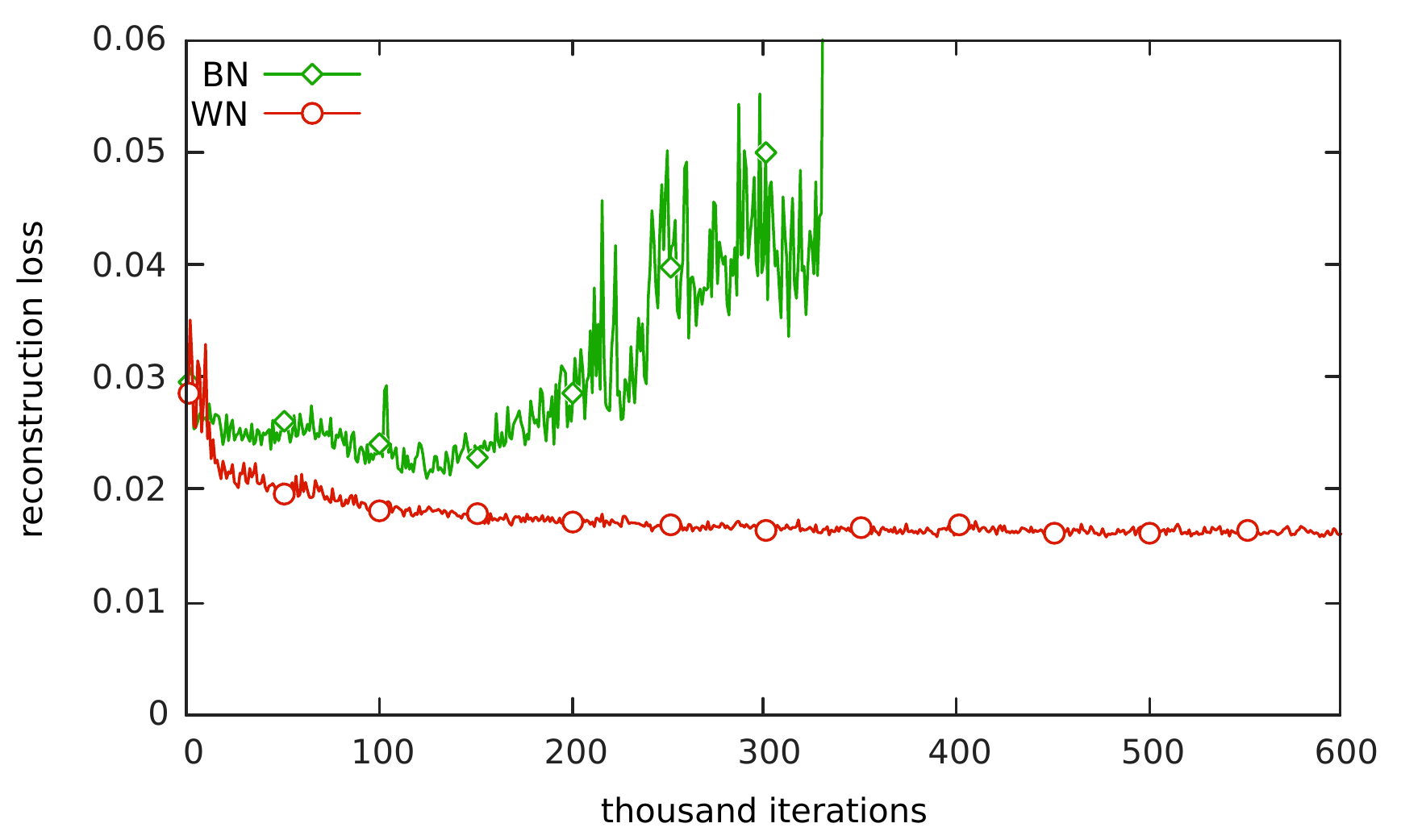}
\end{center}
\caption{\label{fig:lsunrecloss} Reconstruction loss during training on LSUN bedroom dataset.}
\end{figure}

For the vanilla model, training fails constantly, even when we reduce the learning rate by a factor of 10 (to $10^{-5}$), so only the BN and WN models are compared here. The BN model collapsed at iteration 330,800. The WN model was trained with  600,000 iterations. The reconstruction loss is shown in Figure \ref{fig:lsunrecloss} and Table \ref{tab:lsunrecloss}.

\begin{table}
\caption{\label{tab:lsunrecloss} Reconstruction loss of the models on LSUN bedroom dataset}
\begin{center}
\begin{tabular}{cccc}\hline
Model   & Optimal iteration & Running loss & Final loss \\\hline
BN      & 125,000           & 0.020943     & 0.011051        \\
WN      & 478,000           & 0.016266     & 0.008546        \\\hline
\end{tabular}
\end{center}
\end{table}

\begin{figure}
\begin{center}
\includegraphics[width=\linewidth]{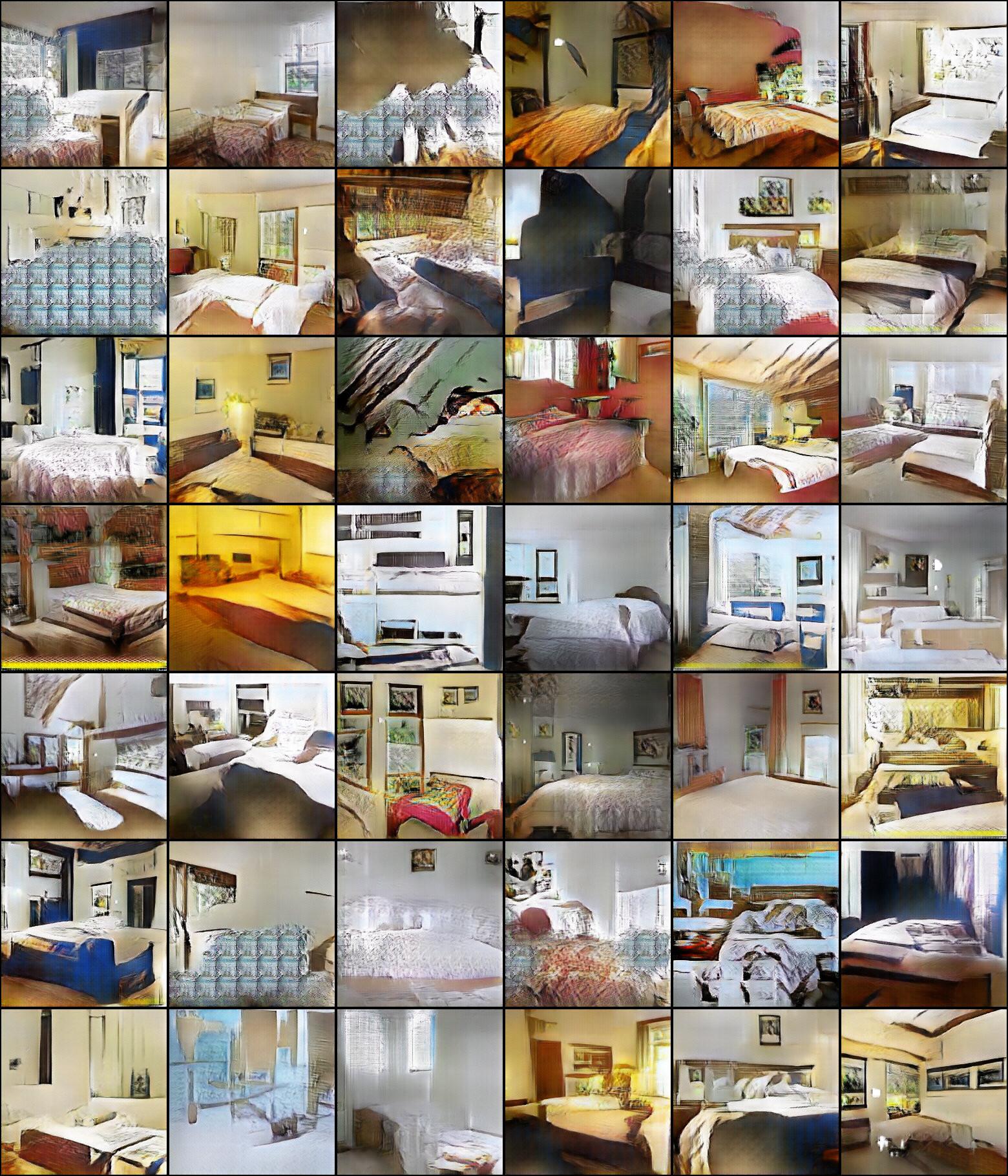}
\end{center}
\caption{\label{fig:lsunbnsamples} LSUN samples generated by the BN model at iteration 125,000}
\end{figure}

\begin{figure}
\begin{center}
\includegraphics[width=\linewidth]{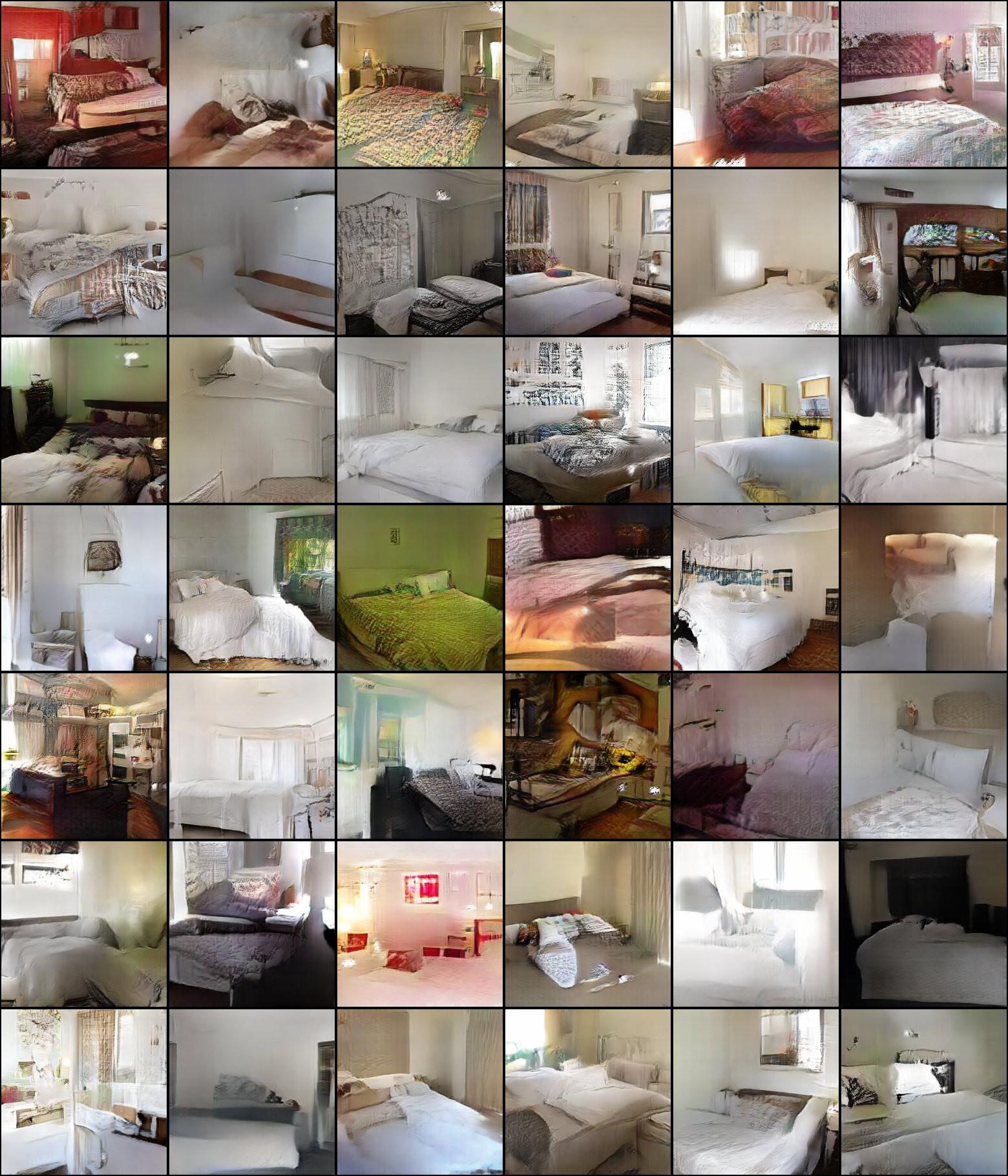}
\end{center}
\caption{\label{fig:lsunwnsamples} LSUN samples generated by the WN model at iteration 299,000}
\end{figure}

\begin{figure}
\begin{center}
\includegraphics[width=\linewidth]{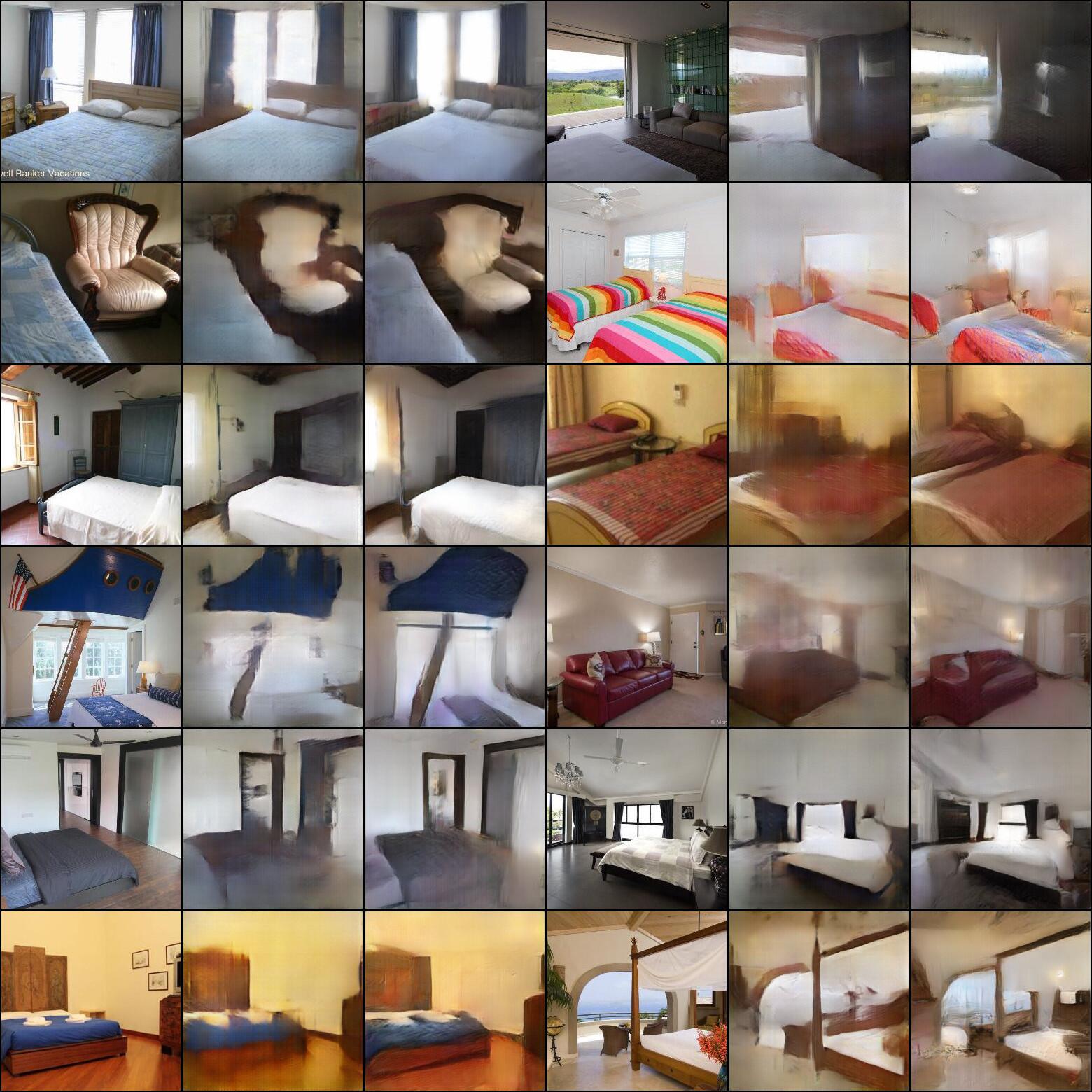}
\end{center}
\caption{\label{fig:lsunreconstructions} LSUN reconstructions. Each group, left to right: test sample, BN model, WN model.}
\end{figure}

Random samples generated by the two models are shown in Figures \ref{fig:lsunbnsamples} and \ref{fig:lsunwnsamples}. Reconstruction of random samples are shown in figure \ref{fig:lsunreconstructions}.

In the BN samples, recurring tile-like artifacts are observed. The best quality samples that are generated by the BN model are arguably sharper and cleaner, while the WN model reproduces details more accurately.

\section{Connection to Wasserstein GAN}
\label{sec:wgan}

For Wasserstein GANs \citep{arjovsky2017wasserstein}, the discriminator is replaced with a critic, that is $K$-Lipschitz-continuous for some constant $K$ and only depends on the structure of the network. To achieve this, they clipped the parameters of the critic network to a small window $[-0.01, 0.01]$ after each parameter update during training.

We claim that our weight-normalized discriminator is Lipschitz-continuous with a small modification:

\newtheorem{wgan0}{Claim}
\begin{wgan0}
\label{thm:wgan0}
The weight-normalized discriminator proposed in this paper is $K$-Lipschitz-continuous for some constant $K$ if the sigmoid layer is removed and the only affine weight-normalized layer is replaced by a strict weight-normalized layer.
\end{wgan0}

To see this, we first prove the following lemma:

\newtheorem{wgan1}[wgan0]{Lemma}
\begin{wgan1}
For a strict weight-normalized layer

\begin{equation}
y_i=\frac{\mathbf{w_i}^T\mathbf{x}}{||\mathbf{w_i}||}\quad,
\end{equation}

where $\mathbf{x}\in\mathbb{R}^n$, $\mathbf{y}\in\mathbb{R}^m$, $\mathbf{W}\in\mathbb{R}^{n\times m}$ is the weight matrix and $\mathbf{w}_i$ the $i$-th column of $\mathbf{W}$. If the loss function of the network is $L$, then

\begin{equation}
\label{eqn:gradbound}
\sum_{i=1}^n\left|\frac{\partial L}{\partial x_i}\right|\le\sqrt{n}\cdot\sum_{i=1}^m\left|\frac{\partial L}{\partial y_i}\right|
\end{equation}
\end{wgan1}

\begin{proof}
\begin{align*}
\sum_{i=1}^n\left|\frac{\partial L}{\partial x_i}\right|=&\sum_{i=1}^n\sum_{j=1}^m\left(\left|\frac{\partial y_j}{\partial x_i}\right|\cdot\left|\frac{\partial L}{\partial y_j}\right|\right)\\
=&\sum_{j=1}^m\left(\left|\frac{\partial L}{\partial y_j}\right|\sum_{i=1}^n\left|\frac{\partial y_j}{\partial x_i}\right|\right)\\
=&\sum_{j=1}^m\left(\left|\frac{\partial L}{\partial y_j}\right|\sum_{i=1}^n \frac{|w_{ij}|}{||\mathbf{w}_j||}\right)\\
\le&\sum_{j=1}^m\left(\left|\frac{\partial L}{\partial y_j}\right|\cdot\sqrt{n}\right)\\
=&\sqrt{n}\cdot\sum_{j=1}^m\left|\frac{\partial L}{\partial y_j}\right|
\end{align*}
\end{proof}

For a strict weight-normalized convolution layer with $c_I$ input channels and kernel size $k_W\times k_H$, change $\sqrt{n}$ in inequality \ref{eqn:gradbound} to $\sqrt{c_I\cdot k_W\cdot k_H}$.

Note that in our implementation, the learned slope of parametric ReLU layers are clipped to $[0,1]$, so the following becomes obvious:

\newtheorem{wgan2}[wgan0]{Lemma}
\begin{wgan2}
For a TPReLU layer with input $\mathbf{x}$ and output $\mathbf{y}$ in $\mathbb{R}^n$,

\begin{equation}
\sum_{i=1}^n\left|\frac{\partial L}{\partial x_i}\right|\le\sum_{i=1}^n\left|\frac{\partial L}{\partial y_i}\right|
\end{equation}

\end{wgan2}

Now it is easy to see that claim \ref{thm:wgan0} is true since for each layer, the sum of absolute value of gradients grows by at most a constant factor. So with such a modification, our discriminator changes into a WGAN critic.

\end{document}